\documentclass[]{clv3}
\usepackage{xcolor}
\usepackage{lineno}
\usepackage{hyperref}
\usepackage{subfigure}
\usepackage{amssymb}
\usepackage{color}
\usepackage{grffile}
\usepackage{tabularx}
\usepackage{ragged2e}

\newcommand{\eqref}[1]{Formula~ \ref{#1}}
\newcommand{\secref}[1]{\S \ref{#1}}
\newcommand{\figref}[1]{Figure~\ref{#1}}
\newcommand{\tabref}[1]{Table~\ref{#1}}
\newcolumntype{C}{>{\Centering\arraybackslash}X}
\renewcommand{\figheight}{3.5cm}
\newcommand{\figheightplus}{3.5cm}
\newcommand{\allfig}[1]{
\centering
\subfigure[Zipf's Law]
{\includegraphics[height=\figheightplus,keepaspectratio]{figs/#1_Zipf.png}}
\hspace*{1.0cm}
\subfigure[Heaps' Law]
{\includegraphics[height=\figheightplus,keepaspectratio]{figs/#1_Heap.png}}
\subfigure[Ebeling's Method]
{\includegraphics[height=\figheight,keepaspectratio]{figs/#1_chars_scaling.png}}
\hspace*{0.0cm}
\subfigure[Taylor's Law]
{\includegraphics[height=\figheight,keepaspectratio]{figs/#1_taylor.png}}
\hspace*{0.0cm}
\subfigure[Long-Range Correlation]
{\includegraphics[height=\figheight,keepaspectratio]{figs/#1_acf.png}}
}

\newcommand{\tinyfigheight}{2.2cm}

\newcommand{\tinyallfig}[1]{
\centering
\subfigure[Zipf's Law]
{\includegraphics[height=\tinyfigheight]{figs/#1_Zipf.png}}
\hspace*{0.0cm}
\subfigure[Heaps' Law]
{\includegraphics[height=\tinyfigheight]{figs/#1_Heap.png}}
\hspace*{0.0cm}
\subfigure[Ebeling's Method]
{\includegraphics[height=\tinyfigheight]{figs/#1_chars_scaling.png}}
\hspace*{-0.2cm}
\subfigure[Taylor's Law]
{\includegraphics[height=\tinyfigheight]{figs/#1_taylor.png}}
\hspace*{-0.2cm}
\subfigure[Long-Range Correlation]
{\includegraphics[height=\tinyfigheight]{figs/#1_acf.png}}
\hspace*{0.0cm}
}

\newcommand{\tinyallfigwordNoEbeling}[1]{
\centering
\subfigure[Zipf's Law]
{\includegraphics[height=\tinyfigheight]{figs/#1_Zipf.png}}
\hspace*{0.0cm}
\subfigure[Heaps' Law]
{\includegraphics[height=\tinyfigheight]{figs/#1_Heap.png}}
\hspace*{0.0cm}
\subfigure[Taylor's Law]
{\includegraphics[height=\tinyfigheight]{figs/#1_taylor.png}}
\hspace*{-0.2cm}
\subfigure[Long-Range Correlation]
{\includegraphics[height=\tinyfigheight]{figs/#1_acf.png}}
\hspace*{0.0cm}
}

\newcommand{\twofigheight}{4.0cm}
\newcommand{\twofig}[1]{
\centering
\subfigure[Zipf's Law]
{\includegraphics[height=\twofigheight]{figs/#1_Zipf.png}}
\subfigure[Taylor's Law]
{\includegraphics[height=\twofigheight]{figs/#1_taylor.png}}
}

\definecolor{darkblue}{rgb}{0, 0, 0.5}
\hypersetup{colorlinks=true,citecolor=darkblue, linkcolor=darkblue, urlcolor=darkblue}

\title{Evaluating Computational Language Models with Scaling Properties of Natural Language}
\runningtitle{Computational Linguistics}
\runningauthor{Takahashi and Tanaka-Ishii}

\begin{document}
\author{Shuntaro Takahashi \thanks{4-6-1 Komaba, Meguro-ku, Tokyo 153-8904, Japan.
} E-mail: takahashi@cl.rcast.u-tokyo.ac.jp}
\affil{Department of Advanced Interdisciplinary Studies,\\ Graduate School of Engineering,\\ The University of Tokyo}

\author{Kumiko Tanaka-Ishii $^{\ast}$ E-mail: kumiko@cl.rcast.u-tokyo.ac.jp}
\affil{Research Center for Advanced Science and Technology, \\ The University of Tokyo}

\maketitle

\begin{abstract}
In this article, we evaluate computational models of natural language with respect to the universal statistical behaviors of natural language. Statistical mechanical analyses have revealed that natural language text is characterized by scaling properties, which quantify the global structure in the vocabulary population and the long memory of a text. We study whether five scaling properties (given by Zipf's law, Heaps' law, Ebeling's method, Taylor's law, and long-range correlation analysis) can serve for evaluation of computational models. Specifically, we test $n$-gram language models, a probabilistic context-free grammar (PCFG), language models based on Simon/Pitman-Yor processes, neural language models, and generative adversarial networks (GANs) for text generation. Our analysis reveals that language models based on recurrent neural networks (RNNs) with a gating mechanism (i.e., long short-term memory, LSTM; a gated recurrent unit, GRU; and quasi-recurrent neural networks, QRNNs) are the only computational models that can reproduce the long memory behavior of natural language. Furthermore, through comparison with recently proposed model-based evaluation methods, we find that the exponent of Taylor's law is a good indicator of model quality.
\end{abstract}

\section{Introduction}
\label{sec:intro}
The question of evaluation methods for computational models of natural language is fundamental in language engineering. Aside from human rating, current evaluation methods rely on the probability distribution produced by the model, or on the $n$-gram similarity between the generated text and a corresponding reference written by human experts. The representative metric of the former type is perplexity. Perplexity quantifies the prediction accuracy of a language model and thus requires its probability distribution. The latter category includes the metrics BLEU \cite{Papineni_2002} and ROUGE \cite{Lin_2004}. These evaluation methods compute the $n$-gram co-occurrence between the generated text and a reference. Hence, the above methods are reasonable for cases in which either the probability distribution of the computational model is explicit and comparable or a corresponding reference is given.

The emergence of intractable models such as generative adversarial networks (GANs) for text generation has revealed the limitation of these conventional evaluation methods. Tentative studies \cite{Yu_2017,LinK_2017,Guo_2017,Subramanian_2017,Lu_2018} have sought to generate natural language text in the adversarial learning framework. Because these models do not explicitly output the probability distribution for prediction, they are evaluated by feeding the generated text to other models, such as a neural language model \cite{Fedus_2018} or a probabilistic context-free grammar (PCFG) \cite{Subramanian_2017}. Although those proposals are promising and worth considering, the effectiveness of the methods for evaluation has not been thoroughly investigated. As an alternative to those approaches, in this article, we test evaluation with the scaling properties of natural language text.

The scaling properties of natural language are the universal statistical behaviors observed in natural language text. For example, Zipf's law characterizes the vocabulary population with a power-law function for the rank-frequency distribution. Recent statistical mechanical studies \cite{Ebeling1995,Altmann2009,plos16,acl18,Tanaka-Ishii_2018} revealed another statistical aspect of natural language, long memory. This refers to the way that sequences of characters or words in natural language universally exhibit clustering, bursty behavior. In particular, results using Taylor's law \cite{acl18,Tanaka-Ishii_2018} show that a natural language text has a consistent range for the Taylor exponent, which quantifies the degree of burstiness in the text.

As the results obtained with scaling properties have clear interpretations, they suggest qualitative implications for language models. For example, evaluation with Zipf's law examines whether a model can properly produce infrequent words. Similarly, evaluation with Taylor's law quantifies whether a model can learn the long memory in a natural language text. In this article, we show that, among the computational models, only neural language models based on recurrent neural networks (RNNs) with a gating mechanism can learn and reproduce the long memory  of natural language text. None of the other models can reproduce this behavior. In addition, our study demonstrates the capabilities of the scaling properties for evaluating language models.

The rest of the article is organized as follows. In \secref{sec:prev}, we review the evaluation metrics that have been widely used for tasks in natural language processing. In \secref{scaling}, we introduce the scaling properties of natural language: those given by Zipf's law, Heaps' law, Ebeling's method, Taylor's law, and long-range correlation analysis. We also explain the methods of applying these scaling properties to evaluate computational models. In \secref{sec:model}, we provide a summary of the models of natural language considered in this article. Specifically, our work covers $n$-gram language models, mathematical language models based on the Simon and Pitman-Yor processes, grammatical models, and neural language models. The experimental procedure and settings are explained in \secref{sec:experiment}. In \secref{sec:evalmetrics}, we assess the scaling properties as evaluation metrics and compare them with other metrics using a PCFG and neural language models. In \secref{sec:evalmodels}, we use the scaling properties to evaluate the models of natural language and discuss the implications of the results. \secref{sec:gan} discusses evaluation of GAN models for text generation. Finally, we conclude our work with a summary in \secref{sec:conc}.

Note that we describe all computational models of natural language considered in this article, as introduced in \secref{sec:model}, by the term \textit{language model}. For some readers this might sound inadequate, because some of these models do not actually form a model to predict subsequent words, e.g., a PCFG and the models based on the Simon and Pitman-Yor processes. Because the term \textit{computational models of natural language} is long, however, for the sake of brevity we simply use the term \textit{language models}.

\section{Previous Evaluation Metrics}
\label{sec:prev}
There are two major approaches to evaluate a language model:
\begin{itemize}
\item directly inspecting some subpart of the model, or
\item verifying the output generated by the model. 
\end{itemize}
This section summarizes previous methods of evaluating models from these two viewpoints, with \secref{sec:perplexity} and \secref{sec:blue} corresponding to the first and second approaches, respectively, and \secref{sec:conjunct} considering both. As clarified in \secref{scaling}, our proposal belongs to the second category.

\subsection{Evaluation using probability distribution: perplexity}
\label{sec:perplexity}

A standard evaluation metric for language models such as $n$-gram and neural language models is the perplexity \cite{perplexity}, which is a measure of the prediction accuracy. Given a test sample $x_1,\ldots,x_N$ of length $N$ and a language model that predicts the probability of words, denoted as $q (x_i)$, the perplexity is defined as the number $e$ to the power of the average log probability of the correct prediction for every word:
\begin{equation}
\displaystyle
\textrm{perplexity}=e^{-\frac{1}{N}\sum^{N}_{i=1}\log{q (x_i)}}.
\label{formula:perplexity}
\end{equation}
Perplexity is usually applied to predict the successive token $x_{i}$ given a context of length $k$, namely, $x_{i-k},\ldots,x_{i-1}$. The probability distribution $q (x_i)$ for prediction must be explicit for evaluation with the perplexity. Moreover, to compare models by using the perplexity, the probability distribution must be defined in a comparable manner. For example, $n$-gram language models and neural language models are comparable, as they predict the next word from the context.

Because perplexity is the current standard metric for automatic evaluation of model quality, the other metrics appearing in this article are compared with the perplexity.

\subsection{Evaluation using reference: BLEU/ROUGE}
\label{sec:blue}
Another popular evaluation metric is the $n$-gram co-occurrence-based approach, including BLEU \cite{Papineni_2002} and ROUGE \cite{Lin_2004}. These metrics are widely used in paired-corpus-oriented tasks such as machine translation and automatic summarization. They evaluate by using statistics of the counts of the same $n$-grams appearing in the machine-generated text and a corresponding reference, which is a correct answer written by an expert. 

These approaches only use the output of a model and thus do not require access to any of its internal elements. Because they require the corresponding reference for computing the $n$-gram co-occurrence, however, their utility is limited to paired-corpus tasks.

As intractable models such as GANs for text generation cannot have an explicit reference, the application of BLEU or ROUGE to those models is not trivial. A series of GAN studies \cite{Yu_2017,LinK_2017,Guo_2017,Lu_2018} quantitatively measured the quality of the generated text with BLEU by regarding the whole training dataset as a reference. The validity of this evaluation method remains questionable, as BLEU was designed for comparison between a pair of a machine-generated text and its correct reference. \citet{Zhu_2018} reported that the application of BLEU with this approach does not provide consistent results with different $n$-grams chosen.

\subsection{Evaluation using other language models}
\label{sec:conjunct}
One approach for evaluation without using either a model distribution or a reference is the use of language models, i.e., evaluation of language models by using other language models. \citet{Fedus_2018} proposed to evaluate GAN-generated text with a neural language model trained with the same natural language dataset. This direction is promising, if the language model is a reliable model of natural language. Even with state-of-the-art neural language models, however, the model quality is limited.

The use of a clear, transparent model for evaluation, such as an $n$-gram language model, would also be a possible method. That approach, however, could only measure models of the $n$-gram structures of natural language and would thus be similar to BLEU evaluation. The use of a PCFG is another possible method of evaluation without a reference. A PCFG is constructed using a parsed corpus such as the Penn Treebank (PTB), and the generated text is parsed with the Viterbi algorithm \cite{Forney_1973}. The algorithm computes the log-likelihood of the text. The PCFG is expected to output a small negative log-likelihood for a grammatically correct sentence. As we demonstrate later, however, it is doubtful that a PCFG could meaningfully evaluate the grammaticality of a sentence. 

\section{Scaling Properties of Natural Language for Evaluation}
\label{scaling}
In this section, we explain scaling properties, the statistical properties of natural language text that have a power-law form. One study on the statistics of natural language reported nine scaling laws \cite{altmann16}. Four of them concern word formation and a network structure, which do not directly relate to language modeling. This leaves five scaling properties, which can be categorized into those for the vocabulary population and those for long memory. These properties are characterized by power-law functions, which involve a power exponent. The exponents of the scaling properties have the capability to characterize the degree of each property. They therefore serve to evaluate whether a language model has the same behavior as natural language text. Specifically, given a text generated by a language model, we set two levels of assessment for evaluation:
\begin{description}
  \item[Q1] Does the scaling property hold qualitatively?
  \item[Q2] How does the exponent differ from that of the training data?
\end{description}
As revealed in the following sections, many models fail to satisfy even the first criterion, especially for long memory. For those models that do satisfy Q1, their exponents can be compared with those of the original text.

Hence, we propose the exponents of scaling properties as metrics to evaluate machine-generated text. Consider a power-law relation $y \propto z^{\kappa}$ for points ($y_1$,$z_1$),$\ldots$, ($y_N$,$z_N$). These points ($y_i$,$z_i$) are calculated for any given text. Let $c$ be the coefficient of the power law, and then the exponent $\kappa$ is estimated by the least-squares method: 
\begin{eqnarray}
  \hat{\kappa},\hat{c} &=& \arg \min_{\kappa,c} \varepsilon (\kappa,c), \label{error} \\
\varepsilon (\kappa,c) &\equiv& \sqrt{\sum_{i=1}^N  (\log y_i - \log cz_i^{\kappa})^2/N}.
\end{eqnarray}

The data points are regressed on a log-log scale. The regression method could be a problem if the errors between the data points and the fitting function are not Gaussian distributed. There are other proposed regression methods such as maximum likelihood estimation for Zipf's law \citep{clauset,altmannprx}. In this article, however, because exponents obtained with the least-squares method are effective in distinguishing machine-generated text from natural language text, and because this method has been a conventional standard, we adopt it for estimation.

\begin{figure*}[tb]
\allfig{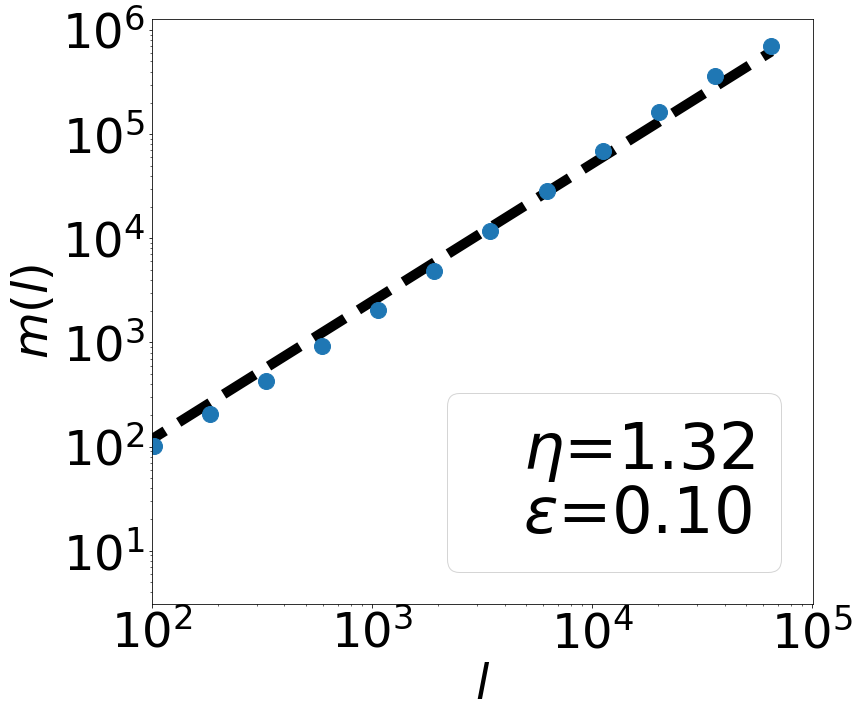}
\caption{Scaling properties of the WT2 dataset. (a) Zipf's law: the rank-frequency distributions of words (red) and word pairs (blue). (b) Heaps' law: the growth of vocabulary size with text length. The solid line is a power-law fitting, and the dashed line represents a power law with exponent $\alpha=1.0$, meaning that all words in a sequence are unique. (c) Ebeling's method: fluctuation analysis of character occurrence. (d) Taylor's law: mean-variance relation of word occurrence. (e) Long-range correlation: temporal correlation of the sequence of the return intervals of rare words. All data points of these five scaling properties are plotted in a log-log scale.}
\label{fig:phy}
\end{figure*}

The following subsections introduce the five scaling properties: Zipf's law, Heaps' law, Ebeling's method, Taylor's law, and a long-range correlation method. As an example, \figref{fig:phy} shows a visual presentation of these methods for the \textit{wikitext-2} (WT2) dataset \cite{Merity_2016}. WT2 is a collected corpus of well-written \textit{Wikipedia} articles, preprocessed by replacing rare words having frequencies under a certain threshold with a meta symbol, \textit{<unk>}. The details of the dataset appear in the first row of \tabref{tab:nldata}, further ahead in \secref{sec:nl}.

\subsection{Vocabulary population}
\label{sec:zipf}

\subsubsection{Zipf's law}
Let $r$ be the rank of a particular word type and $f (r)$ be its frequency. 
The well-known Zipf's law formulates a power-law relation between the frequency and the rank:
\begin{equation}
  f (r) \propto r^{-\alpha}, \label{eq:zipf}
\end{equation}
with $\alpha \approx 1.0$. This scaling behavior generally holds not only for unigrams but also for larger $n$-grams, with smaller exponent values. \figref{fig:phy}(a) shows Zipf distributions for WT2, with unigrams in red and bigrams in blue. Because WT2 replaces rare words, as mentioned before, the tail of the unigram distribution disappears. The Zipf distributions for unigrams and bigrams typically intersect in the middle of the plots. In practice, the plot is not often aligned linearly in a log-log scale, which makes estimation of the exponent $\alpha$ difficult. While previous works have dealt with this problem, it is a sensitive topic and is beyond the scope of this article. We therefore do not estimate $\alpha$ but instead observe the distribution qualitatively.
\subsubsection{Heaps' law}
Heaps' law describes how the vocabulary size grows with the text size following a power-law function. Let $n$ be the length of a text and $v(n)$ be its vocabulary size. Then Heaps' law is formulated as the following relation:
\begin{equation}
 v (n) \propto n^{\beta}, \ \  0 < \beta < 1. \label{eq:heaps}
\end{equation}
\figref{fig:phy}(b) shows the text sizes and corresponding vocabulary sizes for WT2. The exponent $\beta$ is 0.75 with error $\varepsilon=0.13$, which is smaller than $\beta=1.0$ (represented by the dashed black line). There have been multiple debates on how Heaps' law is mathematically related to Zipf's law \citep{BaezaYates_2000,Leijenhorst_2005,Lu_2010}.

\subsection{Long memory}
The statistical mechanics domain has introduced two approaches for quantifying long memory in a time series: fluctuation analysis and the long-range correlation method. We introduce two fluctuation analysis methods, one for characters and one for words, and one long-range correlation method, applied to words. Although these methods are related analytically for a well-formed time series \cite{taylor}, the relation is nontrivial for real phenomena.

\subsubsection{Ebeling's method}
\label{sec:ebeling}
Ebeling's method \cite{Ebeling1995} analyzes the power-law relation between the lengths of subsequences of a text and the variance of the number of characters in the subsequences. Given a set of elements (characters in this method), $W$, let $y (c,l)$ be the counts of character $c$ within subsequences of the text of length $l$. Then, the fluctuation function $m(l)$ is defined as 
\begin{equation}
  m (l) = \sum_{c\in W} m_2 (c,l) \propto l^\eta,  
\end{equation}
where $m_2 (c,l)$ is the variance of the counts $y (c,l)$:
\begin{equation}
  m_2 (c,l) = <y^2 (c,l) > - (< y (c,l)>)^2.
\end{equation}
Theoretically, if a time series is independent and identically distributed (i.i.d.), then $\eta=1.0$, in general, and $\eta >1.0$ if a time series has long memory. \cite{Ebeling1995} reported that the character sequence of the Bible has exponent $\eta = 1.67$, indicating the presence of clustering behavior at the character level. Following the original work, we apply this method at the character level. \figref{fig:phy}(c) shows the fluctuation analysis $m(l)$ for WT2. The exponent is $\eta=1.32$ with error $\varepsilon=0.10$.

\subsubsection{Taylor's law}
\label{sec:taylor}
Taylor's law was originally reported in two pioneering works \cite{taylor-smith,taylor-nature} and has been applied in various domains \cite{taylor}. It describes the power-law relation between the mean and the variance in spatiotemporal observations. In this article, we apply Taylor's law for natural language text as proposed by \citet{acl18,Tanaka-Ishii_2018}.
 
Given a text with a set of words, $W$, for a given segment size $l$ the number of occurrences of a particular word $w \in W$ is counted, and the mean $\mu_w$ and standard deviation $\sigma_w$ are calculated. We thus obtain a scatter plot of $\mu$ and $\sigma$ for all elements of $W$. Taylor's law states the following power-law relation between $\sigma$ and $\mu$ with the Taylor exponent $\zeta$:
\begin{equation}
  \sigma \propto \mu^\zeta \\. 
\end{equation}

\figref{fig:phy}(d) shows the Taylor's law plot for WT2 with $l=5620$ ($l$ can be any value larger than one). The scatter plot generally follows a power-law function with exponent $\zeta=0.62$ and has some deviation from the regression line, with error $\varepsilon=0.15$.

The Taylor exponent takes the range of values $0.50\leq\zeta\leq1.00$, and the two limit values $\zeta=0.50,1.0$ have clear interpretations. For an i.i.d. process, it is proved that $\zeta=0.50$. On the other hand, one case with $\zeta=1.0$ occurs when all segments of length $l$ contain the elements of $W$ with the same proportions. For example, given $W=\{a, b\}$, suppose that $b$ always occurs twice as often as $a$ in all segments (e.g., one segment with three $a$ and six $b$, another segment with one $a$ and two $b$, etc.). Then, both the mean and standard deviation for $b$ are twice those for $a$, and thus $\zeta=1.0$. Therefore, the Taylor exponent quantifies how consistently words co-occur in a text. The Taylor exponent of a natural language text typically has a range of $0.55\leq\zeta\leq0.65$ and never takes $\zeta=0.50$ (which would indicate no long memory). It takes different ranges of values for different types of sequences (e.g., child-directed speech and programming source code). It is therefore expected to have the capability to evaluate machine-generated text.

Ebeling's method and Taylor's law analysis have the following two differences. First, Ebeling's method analyzes the growth of the variance $m(l)$ with respect to the length of the subsequences, $l$, whereas Taylor's law analyzes the variance with respect to the mean frequency within a fixed subsequence length. Second, to acquire an exponent for a text, Ebeling's method takes the sum of the variances over all symbols, whereas Taylor's law obtains the exponent from the individual points for all words.

For the latter reason, Ebeling's method is influenced by a small number of frequently appearing symbols. Because it involves the sum of the variances of all words that follow the power law, the behavior of the exponent $\eta$ often tends to be less sensible than that of the Taylor exponent.

\subsubsection{Long-range correlation}
\label{sec:lrc}
Long-range correlation analysis quantifies the burstiness of word occurrence in a natural language text. The analysis measures the degree of self-similarity within a sequence. Among such analyses, early works proposed mutual-information-based methods \cite{Li_1989,Ebeling1994,Lin_2017}. Such methods compute the mutual information between characters separated by $s$ characters. These works reported that the mutual information decays according to a power law with the distance $s$. \citet{Takahashi_2017} showed, however, that the mutual information method cannot quantify the long-range dependence in word sequences. Moreover, the mutual information between characters decays quickly and reaches a plateau at a distance $s\approx 10^{1}$ for natural language texts such as the collected works of Shakespeare and the PTB dataset.

Another approach to long-range correlation analysis is the use of the autocorrelation function (ACF). The ACF $c(s)$ is defined as the Pearson correlation for two elements of a sequence separated by a distance $s$:

\begin{equation}
	c(s) = \frac{E[(x_{t}-\mu)(x_{t+s}-\mu)]}{\sigma^{2}},
\end{equation}
where $\mu$ and $\sigma$ are the respective mean and standard deviation of the time series $x_{t}$. The value of $c(s)$ ranges between -1 and 1. A time series is said to be long-range correlated if the ACF $c (s)$ for two elements separated by distance $s$ follows a power law:
\begin{equation}
 c (s) \propto s^{-\xi}, \ \ s > 0, 0 < {\xi}<1.
\end{equation}
In the case of application to real-world data, a sequence is said to be long-range correlated if $c (s)$ takes positive values for $s$ until about 1/100 of the length \cite{Lennartz2009}. For sequences without correlation, $c (s)$ fluctuates around zero.

Because the ACF is applicable only for numerical time series, the application of this method for natural language text requires transformation of the sequence of symbols into a numerical time series. Recent methods do so by considering the intervals of word occurrences \cite{plos16}. In this article, we apply a method that measures the ACF of a sequence of the return intervals of rare words, which amounts to $\frac{1}{Q}$ of the text length. With this method, \citet{plos16} reported that power-law decay of the ACF is observed for natural language texts.

\figref{fig:phy}(e) shows the long-range correlation analysis of word sequences in WT2. The hyperparameter was set to $Q=16$ for all results in this article. As seen in the figure, the ACF $c(s)$ always takes positive values up to 1/100 of the sequence length and follows a power-law function (i.e., a straight line in a log-log plot) with exponent $\xi =0.33$ and error $\varepsilon=0.04$. Throughout this article, the error $\varepsilon$ of this metric is only measured for $s \leq 100$.

\begin{table*}[t]
  \scriptsize
  \centering
  \caption{Summary of the datasets  used in this article and their statistics.}
  \label{tab:summary-WT2} \label{tab:nldata}
%

  \begin{tabular}{|l||c|c|c|c|c|c|c|}
    \hline
    &  Tokens & Vocab & \multicolumn{2}{|c|}{Vocabulary Population} & \multicolumn{3}{|c|}{Long Memory} \\ \cline{4-8}
   &  & -ulary& Zipf's & Heaps'  & Ebeling's & Taylor's & Long Range\\
   &&&Law&Law&Method&Law&Correlation\\
   &&&$f(r)\propto r^{-\alpha}$&$v(n)\propto n^\beta$  
   & $m(l) \propto l^{\eta}$&$\sigma \propto \mu^{\zeta}$ & $c(s) \propto s^{-\xi}$\\
    \hline
\multicolumn{8}{|c|}{Wikitext-2 (English, Wikipedia article)}\\\hline
  preprocessed dataset & 2,088,628   & 33,278 & Yes & 0.75 (0.13)& 1.33 (0.10)& 0.62 (0.15) & 0.33 (0.04) \\ \hline
original dataset & 2,088,628 & 76,617  & Yes& 0.78 (0.09)& 1.33 (0.10) & 0.65 (0.11) & 0.32 (0.03) \\
    \hline
    \multicolumn{8}{|c|}{Penn Treebank (English, The Wall Street Journal news article)} \\
    \hline
 preprocessed dataset & 887,521 & 10,000 & Yes & 0.70 (0.16) & 1.23 (0.06) & 0.56 (0.14) & 0.81 (0.24)   \\ \hline
 original dataset & 892,008   & 89,317  & Yes & 0.83 (0.07)& 1.20 (0.05)  & 0.57 (0.06) & 0.60 (0.16) \\   
    \hline
    \multicolumn{8}{|c|}{Shakespeare (old English collection of literature works)} \\
     \hline
   original text  & 740,706 &  83,105 &  Yes &  0.79 (0.07) &  1.24 (0.09) & 0.59 (0.05) & 0.13 (0.02) \\
   \hline
 \multicolumn{8}{|c|}{Hong Lou Meng (Chinese, literature work)} \\
      \hline
original text & 703,034 & 18,312 & Yes  & 0.74 (0.14)   & 1.31 (0.07) & 0.58 (0.07)  & 0.39 (0.04) \\
\hline
   \end{tabular}

\end{table*}

\begin{figure*}[h]
   \allfig{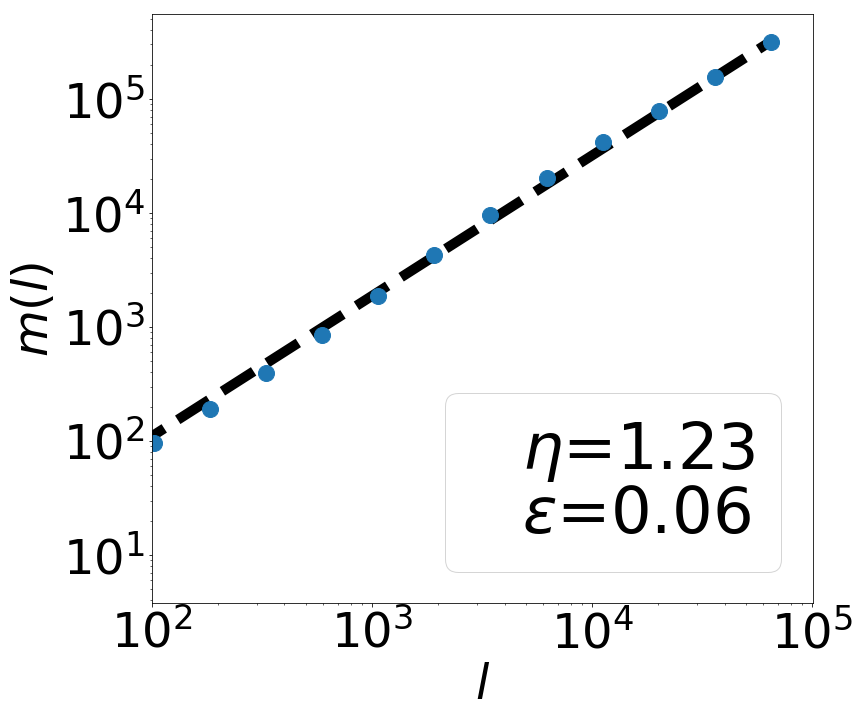}
 \caption{Scaling properties of the Penn Treebank (preprocessed)}
 \label{fig:ptb}
\end{figure*}

\subsection{Examples of scaling properties for other natural language texts}
\label{sec:nl}
Except for Zipf's and Heaps' laws, the scaling properties have hardly appeared in the context of computational linguistics or language engineering. This may be due to the fact that these properties do not directly incorporate semantics or syntax, which are of central concern in those domains. Instead, the properties quantify the universal structures behind natural language in a statistical sense. Those introduced so far are robust and apply to texts across different genres and languages as long as the text is sufficiently long. \figref{fig:ptb} shows the scaling properties of another language modeling dataset, the PTB. This text also satisfies all five scaling properties. They are indeed universal with respect to the genre or even language. More results are shown in Appendix A. \figref{fig:shakespeare} shows the scaling properties of the collected works of Shakespeare, and their exponents are listed in the third block of \tabref{tab:nldata}. Likewise, the scaling properties and exponents for Hong Lou Meng, a Chinese literary work, are shown in \figref{fig:koromu} and listed in the last block of \tabref{tab:nldata}, respectively. Among the exponents, that of the long-range correlation, $\xi$, differs largely among the four datasets considered thus far. In contrast, the other exponents generally take similar values for the datasets.

\section{Computational Models of Natural Language }
\label{sec:model}
This section introduces the computational models of natural language tested in this article. We test four categories of language models: $n$-gram models, grammatical models, language models based on the Simon or Pitman-Yor process, and neural language models. These categories cover the different genres of language models that have appeared in the history of computational linguistics. For every category, some sophisticated, advanced models have been proposed. The experiments reported in \secref{sec:evalmetrics} and \secref{sec:evalmodels}, however, were conducted only with the most recent models whose code was available, except for the $n$-gram models. This served to avoid errors in reimplementation.
\subsection{$N$-gram models}
\label{sec:ngram}
An $n$-gram language model is the most basic model, as it is an $n-1$-ordered Markov model. Let $c (X_1^t)$ be the count of $X_1^t = x_1,x_2,\ldots,x_t$, and then the probability of element $x_{t}$ is calculated as
\begin{equation}
\footnotesize
  P (x_{t+1}|X_{1}^{t}) \approx P (x_{t+1}|X_{t-n+1}^{t}) = 
\frac{c (X_{t-n+1}^{t+1})}{c (X_{t-n+1}^{t})}.
\nonumber
\end{equation}

This article examines $3$-gram and $5$-gram models. Other than the original $n$-gram model, we also test models with a variety of smoothing techniques to improve the perplexity. In particular, linear interpolation \cite{Stolcke_2002}, Katz backoff \cite{Katz_1987}, and Kneser-Ney smoothing \cite{Kneser_1995} have been known to enhance the performance of $n$-gram models. We also set $n=3$ and $n=5$ for these models to compare with the original $n$-gram models. It has been empirically verified that longer context does not necessarily contribute to improving the perplexity and can even degrade performance \cite{Chen_1999}. Simple $n$-gram models, in fact, have been mathematically shown to be incapable of reproducing long memory \cite{Kingman_1963,Lin_2017}.

\subsection{Grammatical models}
The PCFG is a basic grammatical model. We constructed this grammar model with the annotated PTB dataset and used the Natural Language Toolkit (NLTK) \cite{Loper_2002} to generate sentences according to the probabilities assigned to productions. Unlike an $n$-gram model, the PCFG generates a text by using a tree.

\subsection{Language models based on Simon/Pitman-Yor processes}
\label{sec:py}
The Simon and Pitman-Yor processes are abstract models of natural language that reproduce Zipf's law and Heaps' law. These are generative models, and a sequence is formulated over time, either through (1) introduction of new words or (2) sampling from the past sequence. Let $K (X_{1}^{t})$ be the number of word types existing in $X_{1}^{t}$, and let $n_k (X_{1}^{t})$ be the frequency of the $k$th word type in $X_1^t$. The sequence starts with $K (X_{0})=1$ and $X_{0} = x_0$ at $t = 0$. For $t \geq 1$, given a constant $a$ with $0 < a < 1$, the Simon process \cite{simon55} introduces a new word with probability $a$, or a word is sampled from $X_1^t$ with probability $1-a$:
$$
\footnotesize
P (x_{t+1}=w_k) = \left\{
\begin{array}{ll}
     \displaystyle (1-a) \frac{n_k (X_{1}^{t})}{t}  & 1 \leq k \leq K (X_{1}^{t}) \\
        a                           & k = K (X_{1}^{t}) + 1 \\
\end{array}.
\right. 
$$

The Simon process strictly follows Zipf's law with exponent $\alpha=1.0$ and consequently Heaps' law, as well. In contrast, the Pitman-Yor process copes with this problem by decreasing the introduction rate of new words in proportion to $K (X_1^t)$ via another parameter $b$, with $ 0 \leq a < 1$ and $0\leq b$:

$$
\footnotesize
P (x_{t+1} = w_k ) = \left\{
\begin{array}{ll}
     \displaystyle \frac{n_k (X_1^t)-a}{t+b}     &   1 \leq k \leq K (X_1^t) \\
     \displaystyle \frac{a K (X_1^t) + b}{t+b}     & k = K (X_1^t) + 1 \\
  \end{array}.
\right. 
$$

These two parameters serve to produce Zipf's law with slightly convex behavior \cite{goldwater_ml}. The basic models introduced to this point define nothing about how to introduce words: we could simply generate random sequences and examine their scaling properties, because the basic formulations thus far govern the nature of the language models elaborated from these basic models.

By mapping words to the elements produced, we would generate a language model, like the two-stage model proposed in \citet{goldwater_ml}. Here, we consider a more advanced model proposed as the hierarchical Pitman-Yor language model (HPYLM) \cite{teh06}, which integrates the Pitman-Yor process into an $n$-gram model. \footnote{The implementation used in the experiment is available at \url{https://github.com/musyoku/hpylm}. Although HPYLM is an $n$-gram language model and it is possible to calculate the perplexity, the resulting value is not comparable with those of other $n$-gram language models and neural language models. Specifically, the training data requires using \textit{<BOS>} and \textit{<EOS>} to signify the beginning and end of a sentence, respectively. This decreases the perplexity because of the regularities introduced by these insertions, such as \textit{<EOS>} being almost always followed by \textit{<BOS>}.}
\label{sec:py}


\subsection{Neural language models}
\label{sec:neural}
State-of-the-art neural language models are known to outperform $n$-gram language models by the measure of perplexity. The majority of promising neural language models \cite{Mikolov_2012,Melis_2017,Merity_2017,Yang_2017} adopts RNNs. An RNN computes a hidden state $\mathbf{h}_{t}$ recursively from the input $\mathbf{x}_{t}$ and the previous hidden state $\mathbf{h}_{t-1}$ to create an effective flow of past information:
\begin{equation}
	\mathbf{h}_{t} = \Phi (\mathbf{x}_{t},\mathbf{h}_{t-1}).
\end{equation}
The function $\Phi$ depends on the recurrent architecture of the network. A simple RNN model computes the hidden vector $\mathbf{h}_{t}$ as a matrix transformation of $\mathbf{x}_{t}$ and $\mathbf{h}_{t-1}$ by the parameters $\mathbf{W}_{xh}$ and $\mathbf{W}_{hh}$ and a nonlinear transformation by the sigmoid function:
\begin{eqnarray}
	\mathbf{h}_{t} &=& \textnormal{sigmoid}(\mathbf{W}_{xh}\mathbf{x}_{t}+\mathbf{W}_{hh}\mathbf{h}_{t-1}+\mathbf{b}_{h}).
	\label{eq:simple-rnn}
\end{eqnarray}

In modern applications, RNNs with a gating mechanism, such as long short-term memory (LSTM) \cite{Hochreiter_1997}, a gated recurrent unit (GRU) \cite{Cho_2014}, and quasi-recurrent neural networks (QRNNs) \cite{Bradbury_2017}, are often adopted. The recurrent architectures of these models are defined as follows.
\\[5mm]
{ \par}
\textbf{LSTM}
 \begin{eqnarray}
 	\mathbf{i}_{t}&=&\textnormal{sigmoid}(\mathbf{U}_{i}\mathbf{x}_{t}+\mathbf{W}_{i}\mathbf{h}_{t-1}+\mathbf{b}_{i})\\
 	\mathbf{f}_{t}&=&\textnormal{sigmoid}(\mathbf{U}_{f}\mathbf{x}_{t}+\mathbf{W}_{f}\mathbf{h}_{t-1}+\mathbf{b}_{f})\\
 	\mathbf{o}_{t}&=&\textnormal{sigmoid}(\mathbf{U}_{o}\mathbf{x}_{t}+\mathbf{W}_{o}\mathbf{h}_{t-1}+\mathbf{b}_{o})\\
 	\mathbf{\tilde{c}}_{t}&=&\textnormal{tanh}(\mathbf{U}_{\tilde{c}}\mathbf{x}_{t}+\mathbf{W}_{\tilde{c}}\mathbf{h}_{t-1}+\mathbf{b}_{\tilde{c}})\\
 	\mathbf{c}_{t}&=&\textnormal{sigmoid}(\mathbf{f}_{t}\circ \mathbf{c}_{t-1}+\mathbf{i}_{t}\circ \mathbf{\tilde{c}}_{t})\label{eq:gate-lstm}\\ 
 	\mathbf{h}_{t}&=&\textnormal{tanh}(\mathbf{c}_{t})\circ \mathbf{o}_{t}
 \end{eqnarray}
 
\textbf{GRU}
 \begin{eqnarray}
 	\mathbf{r}_{t}&=&\textnormal{sigmoid}(\mathbf{U}_{r}\mathbf{x}_{t}+\mathbf{W}_{r}\mathbf{h}_{t-1}+\mathbf{b}_{r})\\
 	\mathbf{u}_{t}&=&\textnormal{sigmoid}(\mathbf{U}_{u}\mathbf{x}_{t}+\mathbf{W}_{u}\mathbf{h}_{t-1}+\mathbf{b}_{f})\\
 	\tilde{\mathbf{h}}_{t}&=&\textnormal{tanh}(\mathbf{W}_{x}\mathbf{r}_{t}\circ \mathbf{x}_{t}+\mathbf{W}_{h}\mathbf{h}_{t}+\mathbf{b})\\
 	\mathbf{h}_{t}&=&(1-\mathbf{u}_{t})\circ\mathbf{h}_{t-1}+\mathbf{u}_{t}\circ \tilde{\mathbf{h}}_{t} \label{eq:gate-gru}
 \end{eqnarray}

\textbf{QRNNs}
 \begin{eqnarray}
 	\mathbf{z}_{t}&=&\textnormal{sigmoid}(\mathbf{W}^{1}_{z}\mathbf{x}_{t-1}+\mathbf{W}^{2}_{z}\mathbf{x}_{t})\\
 	\mathbf{f}_{t}&=&\textnormal{sigmoid}(\mathbf{W}^{1}_{f}\mathbf{x}_{t-1}+\mathbf{W}^{2}_{f}\mathbf{x}_{t})\\
 	\mathbf{h}_{t}&=&\mathbf{f}_{t}\circ\mathbf{h}_{t-1}+(1-\mathbf{f}_{t})\circ\mathbf{z}_{t} \label{eq:gate-qrnn}
 \end{eqnarray}
Here, the operator $\circ$ denotes element-wise multiplication, the capital symbols $\mathbf{W}$ and $\mathbf{U}$ with subscripts are matrices, and the lowercase symbols $\mathbf{b}$ with subscripts are bias vectors. All these architectures have a gating mechanism (\eqref{eq:gate-lstm}, \eqref{eq:gate-gru}, and \eqref{eq:gate-qrnn}), which balances the use of the states at the previous and current time steps.
 
In this article, we consider a total of nine neural language models. Three of them are based on a simple RNN, a GRU \cite{Cho_2014}, and QRNNs \cite{Bradbury_2017,Merity_2018}. The rest are LSTM-based language models. The first LSTM model is trained without regularizations such as dropout. The second model is AWD-LSTM \cite{Merity_2017}, which applies regularization effectively to achieve competitive prediction performance.
The other four models integrate extended architectures of RNN language models, namely, continuous cache \cite{Grave_2016} and mixture of softmaxes (MoS) \cite{Yang_2017}. Continuous cache is a memory augmentation architecture that computes a cache probability $p_{cache}$ from the $l$ most recent context. It computes the similarity between $\mathbf{h}_{t}$ and $\mathbf{h}_{i}$ to estimate the reappearance of the word at time step $i$. The output probability of the model with continuous cache, denoted as the AWD-LSTM-Cache model, is a linear interpolation of the AWD-LSTM output and the cache probability. We also test a model incorporating the Simon process, denoted as the AWD-LSTM-Simon model. It behaves as a uniform sampling from the past generated sequence and is a special case of AWD-LSTM-Cache. In addition, the MoS architecture reformulates the language modeling task as matrix factorization and is a state-of-the-art language model integrated with AWD-LSTM as the AWD-LSTM-MoS model. Finally, we also consider a combination of all these architectures, denoted as the AWD-LSTM-MoS-Cache model.

The hyperparameters used in our experiments followed the instructions in \cite{Merity_2017,Yang_2017}. The context length (or the length of back-propagation through time) was 70, as given in the references, for both character- and word-based models. The cache window size of the AWD-LSTM-Simon model was set to $10,000$, to balance a large window size with computational efficiency. All the language models were trained to minimize the negative log-likelihood of the training data by stochastic gradient algorithms. Note that the perplexity scores for character- and word-based models are not directly comparable, as they indicate bits per character and per word, respectively.

\section{Experiments}
\label{sec:experiment}
For every language model, a sample text of one million words was generated and evaluated using the metrics explained thus far. We expected models that learned a natural language text to be able to generate a sample text with scaling properties resembling those of the original text. In particular, we expected that the exponent values would be close to those of the original dataset.

The subsequent two sections, \secref{sec:evalmetrics} and \secref{sec:evalmodels}, proceed by examining the scaling properties as applied to models that learned WT2 or the PTB. As introduced in \secref{sec:nl}, these are two standard datasets used as language model benchmarks. For both WT2 and the PTB, the dataset was preprocessed to reduce the vocabulary size. Infrequent words were replaced with \textit{<unk>}, and numbers were replaced with \textit{N} in the PTB \cite{Mikolov_2010}. Language models were then constructed by training with either WT2 or the PTB, except for the Simon and Pitman-Yor processes (but not HPYLM, which does learn) and the PCFG. The PCFG could be constructed only with the PTB dataset, because it requires a parsed corpus, which does not exist for WT2.

\begin{table*}[t]
  \scriptsize
  \centering
  \caption{Summary of the scaling properties of the language models with WT2. $\dagger$ The perplexity measure for HPYLM is not equivalent to that for the $N$-gram and neural language models because of the preprocessing difference. $\ddagger$ The values for these models are in bits per character.}
  \label{tab:summary-WT2}
  
  \begin{tabular}{|l||c|c|c|c|c|c|}
    \hline
    &  Perplexity &  \multicolumn{2}{|c|}{Vocabulary Population} & \multicolumn{3}{|c|}{Long Memory} \\ \cline{3-7}
   &   & Zipf's & Heaps'  & Ebeling's & Taylor's & Long Range\\
   &&Law&Law&Method&Law&Correlation\\
   &&$f(r)\propto r^{-\alpha}$&$v(n)\propto n^\beta$  
   & $m(l) \propto l^{\eta}$&$\sigma \propto \mu^{\zeta}$ & $c(s) \propto s^{-\xi}$\\
    \hline
         \multicolumn{7}{|c|}{Original Dataset} \\
    \hline
    Wikitext-2 (Preprocessed) & -  & Yes &  0.75 (0.13) & 1.32 (0.10)  & 0.62 (0.15)  & 0.33 (0.04) \\
    Wikitext-2 (Original) & -  &Yes &  0.78 (0.09) & 1.33 (0.10)  & 0.65 (0.11)  & 0.32 (0.03)\\
    \hline
    \multicolumn{7}{|c|}{Shuffled Dataset} \\
    \hline
    Wikitext-2(1-gram) & -  & Yes &  0.75 (0.16) & 1.00 (0.01)  & 0.50 (0.02)  & No \\
    Wikitext-2(2-gram) & - &  Yes &  0.76 (0.16) & 1.00 (0.00)  & 0.50 (0.01)  & No \\
    Wikitext-2(5-gram) & - &  Yes &  0.76 (0.16) & 1.00 (0.00)  & 0.50 (0.02)  & No \\
    Wikitext-2(10-gram) & - &  Yes &  0.76 (0.16) & 1.00 (0.00)  & 0.50 (0.02)  & No \\
    \hline

    \multicolumn{7}{|c|}{$N$-gram Language Model} \\
     \hline
   3-gram  & 837.58 &  Yes &  0.79 (0.13) &  1.00 (0.00) & 0.50 (0.02) & No \\
   5-gram &  534.98 &  Yes &   0.78 (0.13)  &  1.00 (0.00) & 0.50 (0.02) & No \\
   linear interpolation & 294.72 & Yes & 0.78 (0.13) & 1.00 (0.00) & 0.50 (0.02) & No  \\
   Katz backoff 3-gram & 285.14 &  Yes & 0.78 (0.13) & 1.00 (0.00) & 0.50 (0.02) & No  \\
   Katz backoff 5-gram & 357.94 &  Yes & 0.78 (0.13) & 1.00 (0.00) & 0.50 (0.02) & No  \\
   Kneser-Ney 3-gram & 204.15 &  Yes & 0.78 (0.13) & 1.00 (0.00) & 0.50 (0.02) & No  \\
   Kneser-Ney 5-gram & 215.44 &  Yes & 0.78 (0.13) & 1.00 (0.00) & 0.50 (0.02) & No  \\
   \hline
 \multicolumn{7}{|c|}{Simon/Pitman-Yor Process and Related Language Model} \\
      \hline
      Simon & - & Yes  & 0.95 (0.15)   &  - & 0.50 (0.01)  & 0.09 (0.03) \\
      Pitman-Yor  & - &  Yes & 0.78(0.09)  &  - & 0.50 (0.01) & No \\
      HPYLM & ($184.34^{\dagger}$) & Yes & 0.78 (0.13) & 1.00 (0.00) & 0.50 (0.02) & No  \\
 \hline
    \multicolumn{7}{|c|}{Neural Language Model (character based)} \\
\hline
LSTM (no regularization)  & ($1.44^{\ddagger}$) & Yes & 0.74 (0.17) & 1.06 (0.05)  & 0.50 (0.01)  & No \\
AWD-LSTM  & ($1.22^{\ddagger}$) & Yes & 0.73 (0.15) & 1.27 (0.10)  & {\bf 0.54} (0.04)  & 0.30 (0.05) \\\hline
\multicolumn{7}{|c|}{Neural Language Model (word based)} \\
    \hline
  Simple RNN  & 164.51 & Yes & 0.79 (0.12) & 1.01 (0.00)  & 0.50 (0.02)  & No \\
 GRU     & 96.22 & Yes & 0.79 (0.11) & 1.12 (0.06)  & {\bf 0.52} (0.03)  & 0.52 (Weak) \\
 QRNN    & 74.74 & Yes & 0.79 (0.11) & 1.08 (0.03)  & {\bf 0.52} (0.03) & 0.57 (0.08) \\
 LSTM (no regularization)&  113.18 &  Yes & 0.78 (0.12) & 1.10 (0.03)   & {\bf 0.52} (0.03)  & 0.43 (0.15) \\
 AWD-LSTM           & 64.27 &  Yes & 0.76 (0.13) & 1.30 (0.15)  & {\bf 0.58} (0.06)  & 0.05 (0.01) \\ 
 AWD-LSTM-Simon     & 61.59 &  Yes & 0.77 (0.10) & 1.25 (0.15)  & {\bf 0.55} (0.05) & 0.03 (0.01) \\
 AWD-LSTM-MoS       & 62.44 & Yes & 0.78 (0.12) & 1.16 (0.07)  & {\bf 0.54} (0.04)  & 0.33 (0.07) \\
 AWD-LSTM-MoS-Cache & 59.21 & Yes & 0.78 (0.11) & 1.20 (0.07)  & {\bf 0.57} (0.07) & 0.29 (0.05) \\
 AWD-LSTM-Cache     & 50.39 & Yes & 0.78 (0.11) & 1.25 (0.10)  & {\bf 0.59} (0.07)  & 0.14 (0.04) \\
 \hline
   \end{tabular}

\end{table*}

\begin{table*}[t]
\scriptsize
  \centering
 \caption{Summary of the scaling properties of the language models with the PTB. $\dagger$ The perplexity measure for HPYLM is not equivalent to that for the $N$-gram and neural language models because of the preprocessing difference. $\ddagger$ The values for these models are in bits per character.}
  \label{tab:summary-PTB}
    \begin{tabular}{|l||c|c|c|c|c|c|}
    \hline
    &  Perplexity  & \multicolumn{2}{|c|}{Vocabulary Population} & \multicolumn{3}{|c|}{
      Long Memory} \\ \cline{3-7}
   &   & Zipf's & Heaps's  & Ebeling's & Taylor's & Long Range\\
&&Law&Law&Method&Law&Correlation\\
   &         & $f(r)\propto r^{-\alpha}$ & $v(n) \propto n^\beta$  
   & $m(l) \propto l^{\eta}$ & $\sigma \propto \mu^{\zeta}$ & $c(s) \propto s^{-\xi}$\\
  \hline
    \multicolumn{7}{|c|}{Original Dataset} \\
  \hline
  Penn Treebank (Preprocessed) & - & Yes &  0.70 (0.16) & 1.23 (0.06)  & 0.56 (0.14)  & 0.81 (0.24) \\
  Penn Treebank (Original)  & -  & Yes &  0.83 (0.07) & 1.20 (0.05)  & 0.57 (0.06)  & 0.60 (0.16) \\
  \hline
  \multicolumn{7}{|c|}{Shuffled Dataset} \\
  \hline
  Penn Treebank (1gram)& - & Yes &  0.72 (0.18) & 1.00 (0.00)  & 0.50 (0.02)  & No \\
  Penn Treebank (2gram) & -  & Yes &  0.72 (0.18) & 1.00 (0.00)  & 0.50 (0.02)  & No \\
  Penn Treebank (5gram) & - & Yes &  0.72 (0.18) & 1.00 (0.00)  & 0.50 (0.02)  & No \\
  Penn Treebank (10gram) & - & Yes &  0.72 (0.18) & 1.00 (0.01)  & 0.50 (0.02)  & No \\
  \hline
  \multicolumn{7}{|c|}{$N$-gram Language Model} \\
  \hline    
  3-gram  & 367.79 &  Yes &   0.71 (0.19) &  0.99 (0.01) & 0.50 (0.02) & No \\
  5-gram  & 561.65 &  Yes & 0.72 (0.21) &  1.00 (0.00)  & 0.50 (0.02) & No \\
  linear interpolation & 238.59  & Yes & 0.71 (0.20) & 1.00 (0.00) & 0.50 (0.02) & No  \\
  Katz backoff 3-gram & 195.65  & Yes & 0.71 (0.19) & 1.00 (0.00) & 0.50 (0.02) & No  \\
  Katz backoff 5-gram & 250.18 & Yes & 0.71 (0.19) & 1.00 (0.00) & 0.50 (0.02) & No  \\
  Kneser-Ney 3-gram & 150.64	& Yes & 0.72 (0.21) & 1.00 (0.00) & 0.50 (0.02) & No  \\
  Kneser-Ney 5-gram & 156.70 & Yes & 0.71 (0.20) & 1.00 (0.00) & 0.50 (0.02) & No  \\

   \hline
 \multicolumn{7}{|c|}{Simon/Pitman-Yor Process and Related Language Model} \\
      \hline
  HPYLM & ($140.49^{\dagger}$) & Yes & 0.73 (0.21) & 1.00 (0.00) & 0.50 (0.02) & No  \\
  \hline
      \multicolumn{7}{|c|}{Grammatical Model} \\
 \hline
   PCFG   &  - &   Yes &  0.73 (0.19) & 1.00 (0.00) & 0.50 (0.02) & No \\
   \hline
    \multicolumn{7}{|c|}{Neural Language Model (character based)} \\
    \hline
LSTM (no regularization)  & ($1.38^{\ddagger}$)  & Yes & 0.79 (0.08) & 1.03 (0.01)  & 0.50 (0.01)  & No \\
    AWD-LSTM  & ($1.18^{\ddagger}$)  & Yes & 0.76 (0.12) & 1.10 (0.03)  & {\bf 0.51} (0.02)  & 0.40 (0.10) \\
\hline
    \multicolumn{7}{|c|}{Neural Language Model (word based)} \\
 \hline
 Simple RNN  & 123.96 & Yes & 0.71 (0.19) & 1.00 (0.01) & 0.50 (0.02)  & 0.74 (Weak) \\
 GRU     & 85.05 & Yes & 0.71 (0.18) & 1.05 (0.02)  & 0.50 (0.02)  & 0.40 (Weak) \\
 QRNN    & 62.65 & Yes & 0.71 (0.18) & 1.10 (0.03)  & {\bf 0.51} (0.02)  & 0.54 (Weak) \\
LSTM (no regularization)&111.79 & Yes & 0.71 (0.19) & 1.04 (0.01) & {\bf 0.51} (0.02)  & 0.84 (Weak) \\
AWD-LSTM           & 56.40 & Yes & 0.71 (0.18) & 1.06 (0.02)  & {\bf 0.51} (0.03)  & 0.69 (Weak) \\ 
AWD-LSTM-Simon     & 57.85 &  Yes & 0.72 (0.16) & 1.04 (0.01)  & {\bf 0.51} (0.03)  & No \\
AWD-LSTM-MoS       & 54.77 & Yes & 0.71 (0.18) & 1.10 (0.03)  & {\bf 0.52} (0.04)  & 0.77 (Weak) \\
AWD-LSTM-MoS-Cache & 54.03 &  Yes & 0.71 (0.18) & 1.13 (0.04) & {\bf 0.55} (0.06) & 0.61 (Weak) \\
AWD-LSTM-Cache     & 52.51 & Yes & 0.72 (0.17) & 1.07 (0.02) & {\bf 0.53} (0.05)  & 0.57 (Weak) \\
\hline
  \end{tabular}

\end{table*}

\tabref{tab:summary-WT2} and \tabref{tab:summary-PTB} list the perplexity and the scaling exponents of the models for the WT2 and PTB datasets, respectively. Each row presents the results for a single text, either real or machine generated. The perplexity is not reported for the Simon model, the Pitman-Yor process, or the PCFG. For the two mathematical models, it was not measured because they do not have references for computing the prediction accuracy. The perplexity of the PCFG is not reported because its computation does not trivially match that of the $n$-gram and neural language models.

The first blocks in each table indicate the properties of the original datasets with and without preprocessing. The second blocks list the results for shuffled datasets, which preserve parts of the $n$-gram structure. They were tested to check the behavior of the evaluation metrics on randomized texts. The shuffled datasets were expected to lose long memory and were largely different from the original natural language texts. The shuffling was conducted as follows. As an example, the text \textit{ABCDEFGHI} was first split into $3$-gram chunks, giving \textit{ABC/DEF/GHI}. Then, the chunks were shuffled randomly to obtain a $3$-gram shuffled dataset (i.e., \textit{DEF/GHI/ABC}). Note that this shuffling does not preserve some $n$-gram structures, such as \textit{BCD} and \textit{FGH}, in the original text. The remaining blocks correspond to the results for the language models introduced. The grammatical model category is absent in \tabref{tab:summary-WT2} because of the lack of a parsed corpus for WT2. Appendix B includes all figures showing the scaling properties.
\section{Evaluation of Metrics}
\label{sec:evalmetrics}
The first columns of \tabref{tab:summary-WT2} and \tabref{tab:summary-PTB} list the perplexities of the language models. The blank symbol ``-'' appears in rows for which the perplexity is not available: the original and shuffled datasets are not language models, while the Simon/Pitman-Yor processes and the grammatical model have different definitions of probability and cannot be measured comparably with the $n$-gram and neural language models. The perplexity scores in parentheses were measured comparably but are not comparable with the other values because of their different implementations of preprocessing, as explained at the ends of \secref{sec:py} and \secref{sec:neural}.

In terms of perplexity, the neural language models consistently outperformed the $n$-gram models. Among the $n$-gram models, Kneser-Ney smoothing consistently outperformed the other smoothing techniques. The $3$-gram models sometimes had better perplexity than the $5$-gram models did, as the training datasets in this experiment were not especially large (see \tabref{tab:nldata}). Among the neural language models, the simple RNN model had the worst perplexity. The RNNs with a gating mechanism improved the perplexity over that of the simple RNN model. In particular, the AWD-LSTM model performed the best among the RNN language models. The additional architectures of the cache mechanism and MoS contributed to improving the perplexity.

\subsection{Metrics of scaling properties}
\label{sec:eval_metric}
The proposed evaluation metrics should be compared with another evaluation metric that is assumed plausible. In this article, the perplexity is adopted as such a metric. As perplexity has been the standard evaluation metric in language modeling and the prediction accuracy is of primary importance for that application, we compare the metrics derived from the scaling properties by comparing them with the perplexity and consider how they correlate with it.

The third to seventh columns of \tabref{tab:summary-WT2} and \tabref{tab:summary-PTB} list the respective results for the scaling properties: Zipf's law, Heaps' law, Ebeling's method, Taylor's law, and the long-range correlation. Even when the perplexity was not computable, the properties could all still be examined regardless of the kind of language model, except for Ebeling's method, because it applies to characters. Overall, except for the long-range correlation, the results were consistent across the datasets: when a scaling law was followed by one dataset, then it was also followed by the other dataset.

All the language models qualitatively satisfied Zipf's law. We indicate this by \textit{Yes} in the tables for the reason stated in \secref{sec:zipf}. Relatedly, all the language models also satisfied Heap's law. These two properties, however, are present even with a unigram language model. Despite their fame, Zipf's law and Heaps' law have no capacity to distinguish randomized and real text. It is therefore not a challenge for language models to satisfy Zipf's and Heaps' laws.

In contrast, the metrics of long memory were capable of quantifying the quality of machine-generated texts. For Ebeling's method (first column of the \textit{Long Memory} vertical block), the exponent of the original dataset was $\eta=1.32$ for WT2 and $\eta=1.23$ for the PTB, whereas that of both shuffled datasets was $\eta=1.00$, thus indicating no long memory in the latter. The neural language models had exponents between $\eta=1.10$ and $\eta=1.30$ for WT2, and between $\eta=1.04$ and $\eta=1.13$ for the PTB, whereas the other language models were the same as i.i.d. behavior. Ebeling's method therefore could verify the text quality to a certain extent.

The last column in each table lists the results for the long-range correlation. If the text was not long-range correlated, this is denoted by \textit{No} or \textit{Weak}: \textit{No} if more than one value was negative for $s\leq10$, or \textit{Weak} if there was one negative value for $s\leq100$. Such arbitrariness of judgement is one disadvantage of this metric. In addition, even though it has good correspondence with the other two metrics of long memory, it has two further disadvantages. First, the exponent has poor correlation with the perplexity. The second disadvantage was exhibited in the degree of long-range correlation listed for the Simon model. The degree was high at the beginning and did not decay (see \figref{fig-appendix:simon} in Appendix B). As the Simon model had more new words later in a sequence, the correlation stayed large even for two sequences with a large distance between them. Therefore, this non-decaying phenomenon was due not to burstiness but to a different characteristic specific to the Simon process. The Taylor exponent for the Simon process was $\zeta=0.50$, indicating that the long-range correlation observed was not due to long memory behavior.

\begin{figure}[t]
  \centering
  \includegraphics[width=1.0\textwidth]{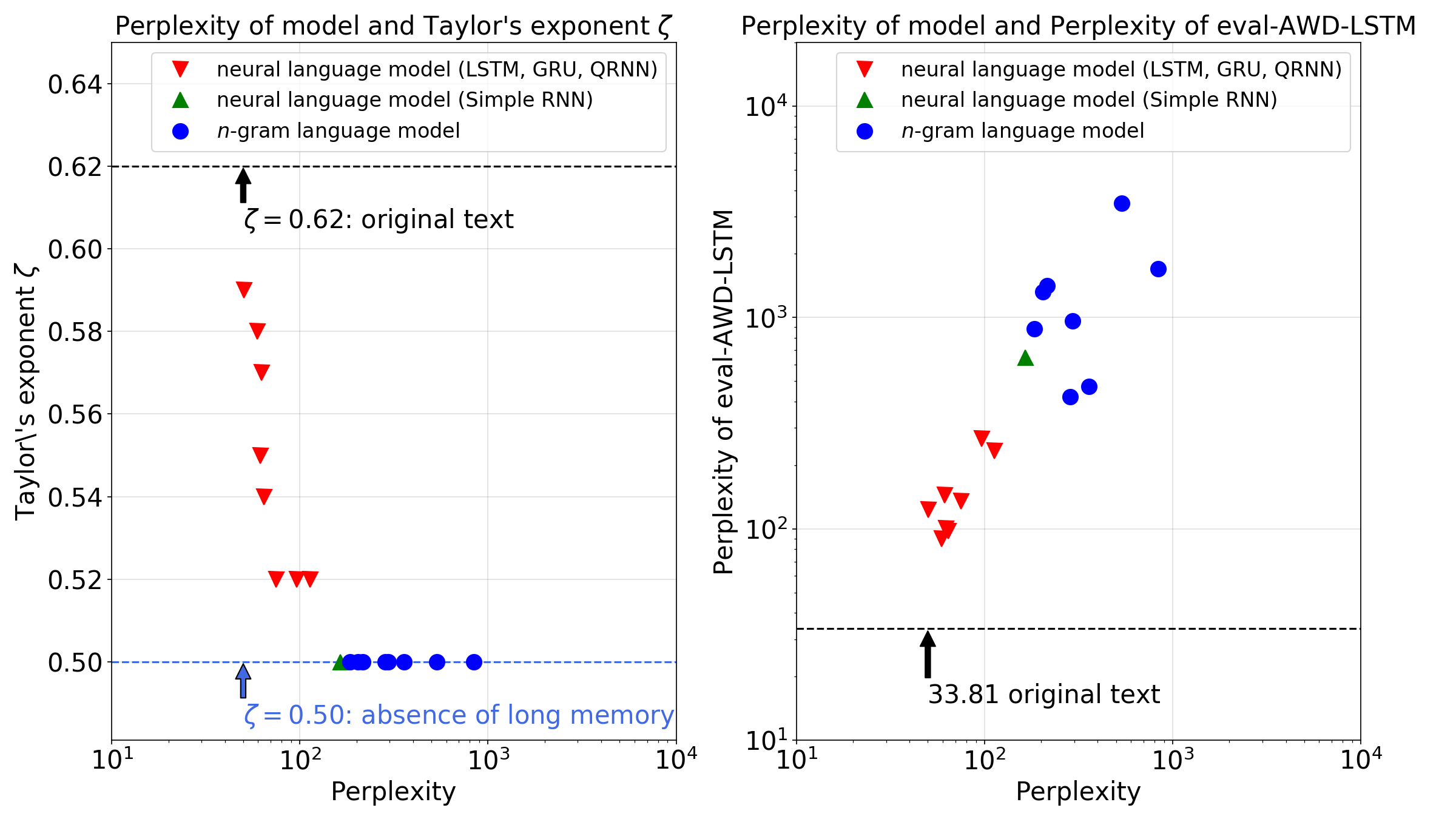}
  \caption{Scatter plots of the perplexity of various models with respect to the Taylor exponent $\zeta$ (left) and the perplexity of the eval-AWD-LSTM model (right) for the WT2 dataset. (Left) The Taylor exponents of the $n$-gram language models were consistently $\zeta=0.50$, which indicates the absence of long memory. In contrast, the neural language models had Taylor exponents of $\zeta>0.50$, which indicates the presence of long memory in the generated texts. (Right) The perplexity of eval-AWD-LSTM had clear, positive correlation with the perplexities of the language models.}
  \label{fig:scatter}
\end{figure}

Finally, the Taylor exponent $\zeta$ seemed the most reliable metric among those derived from the scaling properties. The left panel of \figref{fig:scatter} shows the correlation between the perplexity of the models and the Taylor exponent $\zeta$. As the perplexity decreased, the Taylor exponent $\zeta$ showed a steep increase. Because the exponent quantifies the degree of burstiness of word occurrence, this result indicates that the better models in terms of perplexity can also reproduce that statistical property.

Overall, the scaling properties of long memory serve for evaluation of generated texts. The Taylor exponent $\zeta$ especially has the capability for evaluation.

\begin{figure*}[t]
	\centering
	\includegraphics[width=10.0cm]{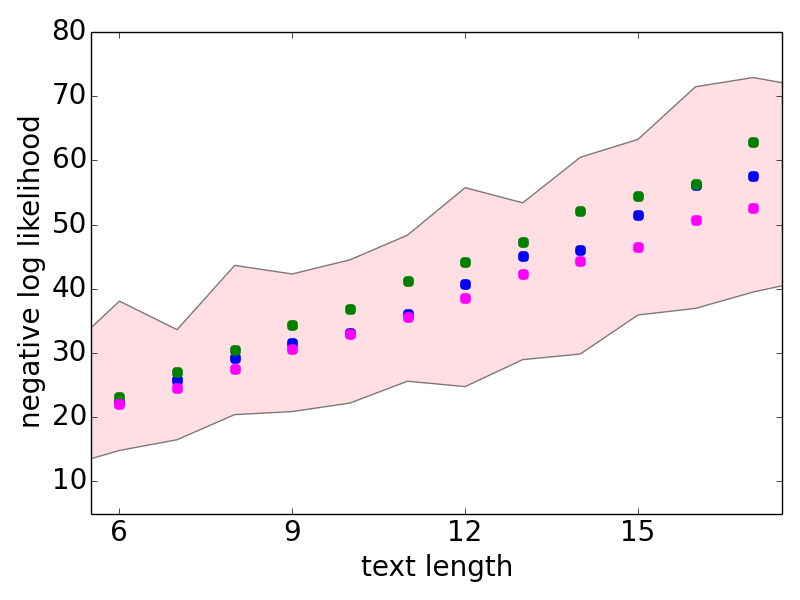}
	\caption{Average negative log-likelihood of a PCFG for different sentence lengths from the PTB dataset (magenta), $n$-word chunks from the AWD-LSTM-Cache model (blue), and 5-grams from the shuffled PTB dataset (green). The area shaded red represents the upper and lower bounds of the negative log-likelihood of the PCFG for the PTB dataset.}
	\label{fig:pcfg-dist}
\end{figure*}

\begin{table}[]
\setlength\extrarowheight{2pt}
 \centering
 \scriptsize
 \caption{Evaluation of language models by using the AWD-LSTM model (trained with WT2), in comparison with using the perplexity and the Taylor exponent.}
 \label{table:eval_WT2}

\begin{tabular}{|l||c|c|c|}
\hline
 & \multicolumn{1}{l|}{Perplexity} & \multicolumn{1}{l|}{Taylor exponent}  & \multicolumn{1}{p{1.5cm}|}{Perplexity from eval-AWD-LSTM} \\ \hline
\multicolumn{4}{|c|}{Original Dataset}  \\ \hline
Wikitext-2 (Preprocessed)   & -  & 0.62 (0.15) &  33.81\\ \hline
\multicolumn{4}{|c|}{Shuffled Dataset}  \\ \hline
Wikitext-2 (1-gram) & -  & 0.50 (0.02) &  7389.15 \\ \hline
Wikitext-2 (2-gram) & -  & 0.50 (0.02) & 2405.15 \\ \hline
Wikitext-2 (5-gram) & -  & 0.50 (0.02) & 559.92  \\ \hline
Wikitext-2 (10-gram)& -  & 0.50 (0.02) & 236.49  \\ \hline
\multicolumn{4}{|c|}{$N$-gram Language Model}   \\ \hline
3-gram  & 837.58 & 0.50 (0.02)  & 3730.74 \\ \hline
5-gram   & 534.98 & 0.50 (0.02) & 7532.91 \\ \hline
linear interpolation  & 294.72 & 0.50 (0.02) & 1371.75 \\ \hline
Katz backoff 3-gram & 285.14 & 0.50 (0.02) & 663.74  \\ \hline
Katz backoff 5-gram & 357.94 & 0.50 (0.02) & 664.25 \\ \hline
Kneser-Ney 3-gram & 204.15 & 0.50 (0.02) & 2562.24\\ \hline
Kneser-Ney 5-gram & 215.44 & 0.50 (0.02) & 2743.65   \\ \hline
HPYLM & 184.34 & 0.50 (0.02) & 884.76   \\ \hline
\multicolumn{4}{|c|}{Neural Language Model} \\ \hline
Simple RNN  & 164.51  & 0.50 (0.02) & 645.64  \\ \hline
GRU  & 96.22  & 0.52 (0.03) & 266.33  \\ \hline
QRNN  & 74.74  & 0.52 (0.03) & 135.68  \\ \hline
LSTM (no regularization) & 113.18 & 0.52 (0.03) & 177.12 \\ \hline
AWD-LSTM & 64.27  & 0.58 (0.06) & 88.73 \\ \hline
AWD-LSTM-Simon & 61.59  & 0.55 (0.05) &  130.52\\ \hline
AWD-LSTM-MoS  & 62.44  & 0.54 (0.04) & 97.89  \\ \hline
AWD-LSTM-MoS-Cache  & 59.21  & 0.57 (0.07) & 164.39  \\ \hline
AWD-LSTM-Cache  & 50.39  & 0.59 (0.07) & 109.02  \\ \hline
\end{tabular}

 \end{table}

\begin{table}[]
\setlength\extrarowheight{2pt}
\centering
\scriptsize
\caption{Evaluation of language models by using the AWD-LSTM model (trained with the PTB), in comparison with using the perplexity and the Taylor exponent.}
\label{table:eval_PTB}

\begin{tabular}{|l||c|c|c|c|}
\hline
 & \multicolumn{1}{l|}{Perplexity} & \multicolumn{1}{l|}{Taylor exponent} & \multicolumn{1}{p{1.5cm}|}{Perplexity from eval-AWD-LSTM} \\ \hline
\multicolumn{4}{|c|}{Original Dataset} \\ \hline
Penn Tree Bank (Preprocessed)   & -  & 0.56 (0.14)  & 40.70 \\ \hline
\multicolumn{4}{|c|}{Shuffled Dataset} \\ \hline
Penn Tree Bank (1-gram)& -  & 0.50 (0.02)    & 3698.52  \\ \hline
Penn Tree Bank (2-gram)& -  & 0.50 (0.02)    &  1328.39 \\ \hline
Penn Tree Bank (5-gram)& -  & 0.50 (0.02)   & 351.22  \\ \hline
Penn Tree Bank (10-gram)   & -  & 0.50 (0.02)   & 166.93  \\ \hline
\multicolumn{4}{|c|}{$N$-gram Language Model}   \\ \hline
3-gram   & 367.79  & 0.50 (0.02) & 1697.99 \\ \hline
5-gram& 561.65 & 0.50 (0.02) & 3463.88 \\ \hline
linear interpolation  & 238.59 & 0.50 (0.02)   & 965.58 \\ \hline
Katz backoff 3-gram   & 195.65 &  0.50 (0.02)  & 420.48\\ \hline
Katz backoff 5-gram   & 250.18 &  0.50 (0.02) & 471.03\\ \hline
Kneser-Ney 3-gram & 150.64 &  0.50 (0.02) & 1324.67 \\ \hline
Kneser-Ney 5-gram &156.70 &  0.50 (0.02) & 1411.14 \\ \hline
HPYLM &140.49 &  0.50 (0.02) & 412.13 \\ \hline
\multicolumn{4}{|c|}{Neural Language Model}\\ \hline
Simple RNN  & 123.96  & 0.50 (0.02) &  321.31 \\ \hline
GRU  & 85.05  & 0.50 (0.02) & 258.12  \\ \hline
QRNN  & 62.65  & 0.51 (0.02) & 113.22  \\ \hline
LSTM (no regularization)& 113.18 & 0.51 (0.02) & 234.05 \\ \hline
AWD-LSTM  & 64.27  & 0.51 (0.03) & 90.01 \\ \hline
AWD-LSTM-Simon& 61.59  & 0.51 (0.03) & 144.45  \\ \hline
AWD-LSTM-MoS & 62.44  & 0.52 (0.04) & 97.73 \\ \hline
AWD-LSTM-MoS-Cache & 59.21  & 0.55 (0.06)  & 100.56 \\ \hline
AWD-LSTM-Cache & 50.39  & 0.53 (0.05) & 123.32  \\ \hline
\end{tabular}

\end{table}

\subsection{Comparison with PCFG- and language-model-based evaluation}
\label{sec:experiment-models}
Next, we test the effectiveness of using the negative log-likelihood from a PCFG \cite{Subramanian_2017} and the perplexity obtained from a neural language model \cite{Fedus_2018}. The results show how PCFG-based evaluation is not effective, in contrast to evaluation based on the scaling properties.

In principle, the negative log-likelihood of a PCFG evaluates the grammaticality of text. \citet{Subramanian_2017} used the negative log-likelihood of a PCFG to evaluate GAN-generated texts. The scatter plots in \figref{fig:pcfg-dist} show the average negative log-likelihood from a PCFG for the PTB dataset (magenta), the PTB dataset shuffled with 5-grams (green), and the AWD-LSTM-Cache model (blue). Because the PTB dataset is annotated, the negative log-likelihood was calculated for every sentence, and the values were plotted for different sentence lengths. As for the other two cases, because the outputs had no sentence boundaries indicated in the training data, consecutive parts of a given length $n$ were randomly extracted from the text and fed to the PCFG parser, and the negative log-likelihood was then calculated. The NLTK \cite{Loper_2002} parser implementation was used in this work. The shaded area in red represents the upper and lower bounds of the original PTB dataset.

The average negative log-likelihood of a sentence has a strong linear correlation with its length, and the values for the PTB dataset were consistently lower than those for the generated text of the AWD-LSTM-Cache model and the 5-gram shuffled text. The differences from the original PTB dataset, however, were not significant, even though the 5-gram and AWD-LSTM-Cache results were calculated merely for $n$-word random chunks. Moreover, the average values for the 5-gram shuffled text and the machine-generated text were {\em within} the range of the PTB's upper and lower bounds. This indicates that the negative log-likelihood from a PCFG is probably not usable for evaluating machine-generated texts.

Apart from the PCFG, \citet{Fedus_2018} proposed to evaluate the quality of GAN-generated texts with the perplexity computed from a neural language model. We next test whether that method provides a good measure of the language models considered here. Accordingly, we used the AWD-LSTM model to evaluate the texts generated by the $n$-gram and neural language models. To avoid confusion, we call this the eval-AWD-LSTM model. It was trained with the WT2 and PTB datasets to evaluate the texts generated by the various other models (including AWD-LSTM itself).

The perplexity of eval-AWD-LSTM was calculated for each machine-generated text by formula (\ref{formula:perplexity}). The rightmost columns of \tabref{table:eval_WT2} and \tabref{table:eval_PTB} list the results, and the right panel of \figref{fig:scatter} shows a scatter plot of the perplexity of the models with respect to the perplexity of eval-AWD-LSTM. This method seemed to work well, especially in globally distinguishing the $n$-gram and neural language model categories: the former category had perplexities above 600, whereas the latter category had almost all values below 200 for WT2. The eval-AWD-LSTM perplexity could not, however, detect the differences among the $n$-gram language models nor among the neural language models (e.g., between Katz backoff and Kneser-Ney, or AWD-LSTM and AWD-LSTM-Cache). The bias caused by the evaluation model is also a problem with this method. In the experiment, AWD-LSTM was the best model by eval-AWD-LSTM evaluation for both the WT2 and PTB datasets. It is likely that worse-performing models whose behavior is similar to that of the evaluation model are evaluated more highly than are other models that have higher fluency but behave differently from the evaluation model.

Overall, the evaluation methods using other language models were not consistent. The PCFG-based evaluation could not even clearly distinguish between the shuffled and original datasets. Evaluation based on a neural language model could detect the difference between the $n$-gram and neural language models, but it could not distinguish quality within those categories of language models. Compared with those methods, the Taylor exponent $\zeta$ had a clearer correlation with the perplexity of the models. Specifically, the exponent satisfied $\zeta=0.50$ for all $n$-gram language models. It was larger than $0.50$ only for the neural language models whose perplexity was better than that of the $n$-gram language models. Among the neural language models, the Taylor exponent took high values for the AWD-LSTM family, which had better perplexity than the GRU and QRNN models and the LSTM model without regularization.

\begin{figure*}[t]
  \twofig{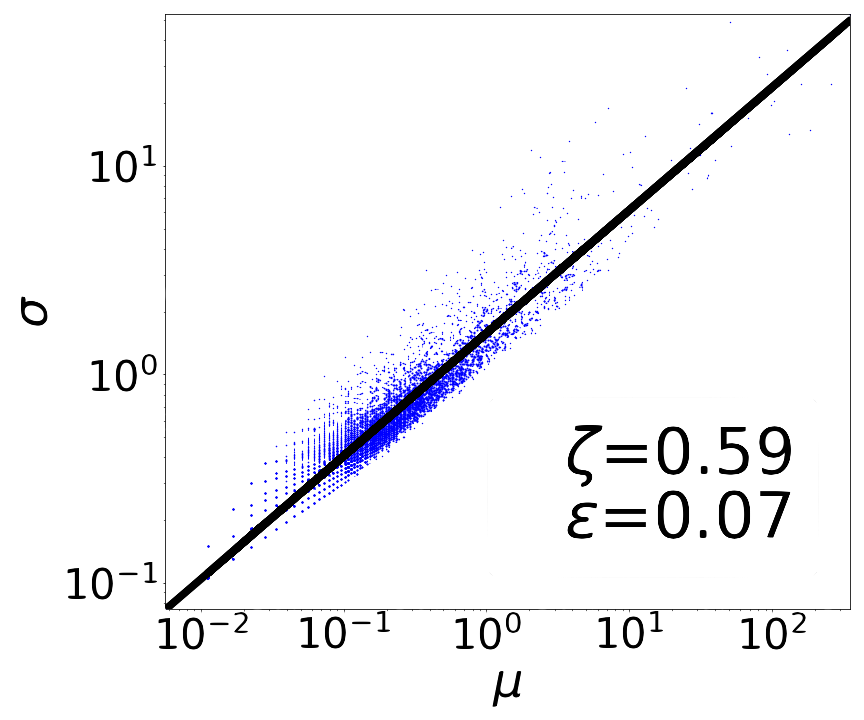}
\caption{Scaling properties of the AWD-LSTM-Cache model trained with WT2. 
\label{fig:awdcache}}
\end{figure*}

\section{Evaluation of Models}
\label{sec:evalmodels}
In this section, we apply the evaluation of metrics in \secref{sec:eval_metric} to discuss the scaling properties of the language models. All language models tested in the experiments satisfied the scaling properties of vocabulary population, Zipf's law and Heaps' law. These properties are relatively easy for models to reproduce, because they concern the static probability distribution of words.

In contrast, many of the language models failed to reproduce long memory behavior. The sole exception was the Simon process, which presented strong long-range correlation, but this was not caused by burstiness, as explained in \secref{sec:eval_metric}. The lack of long memory in $n$-gram language models is supported by an analytical argument about Markov models, as mentioned in \secref{sec:ngram}. The failure of the PCFG model in our experiment setting can be explained by its lack of inter-sentence structure.

Even among the neural language models, the simple RNN model failed to reproduce long memory. The Taylor exponent was $\zeta=0.50$, and the other metrics also indicated that the generated text did not have long-range dependence. In contrast, the RNNs with a gating mechanism (LSTM, GRU, and QRNNs) could reproduce long memory behavior. The Taylor exponents of the GRU and QRNN language models were both $\zeta=0.52$ for WT2, which indicates the presence of long memory to a certain extent. The LSTM language models were consistently the best at reproducing long memory behavior of natural language text for WT2 and the PTB at both the character level and the word level.

\figref{fig:awdcache} shows (a) Zipf's law and (b) Taylor's law results for the AWD-LSTM-Cache model trained with WT2, which was the best performing model in terms of perplexity. \figref{fig:awdcache}(a) demonstrates that the Zipf's law behavior of the dataset shown in \figref{fig:phy}(a) was well recovered. Likewise, \figref{fig:awdcache}(b) demonstrates how well the AWD-LSTM-Cache model captured and reproduced the Taylor's law behavior shown in \figref{fig:phy}(d). While the Taylor exponent for the original dataset was $\zeta=0.62$, the AWD-LSTM-Cache model had a Taylor exponent of $\zeta=0.59$ for WT2. The data points in \figref{fig:phy}(d) were more widely scattered around the regression line than those in \figref{fig:awdcache}(b) were. Even with the well-performing neural language models, however, the scaling properties of long memory were not fully recovered. These differences represent gaps between the natural language text and the language model, which may indicate room for improvement.

\begin{table*}[t]
  \scriptsize
  \centering
  \caption{Summary of statistics for the COCO image dataset.}
  \label{tab:coco}
  %

  \begin{tabular}{|l||c|c|c|c|c|c|c|}
    \hline
    &  Tokens & Vocab & \multicolumn{2}{|c|}{Vocabulary Population} & \multicolumn{3}{|c|}{Long Memory} \\ \cline{4-8}
   &  & -ulary& Zipf's & Heaps'  & Ebeling's & Taylor's & Long Range\\
   &&&Law&Law&Method&Law&Correlation\\
   &&&$f(r)\propto r^{-\alpha}$&$v(n)\propto n^\beta$  
   & $m(l) \propto l^{\eta}$&$\sigma \propto \mu^{\zeta}$ & $c(s) \propto s^{-\xi}$\\
    \hline
    \multicolumn{8}{|c|}{Image COCO (English, collection of image caption)} \\
 \hline
\begin{tabular}[l]{@{}l@{}}original dataset \end{tabular} & 105,933   & 6,095  & Yes & 0.76 (0.09) &  0.99 (0.03)  & 0.50 (0.04)                                                & No \\ \hline
   \end{tabular}

\end{table*}
  
\begin{figure}[h]
	\twofig{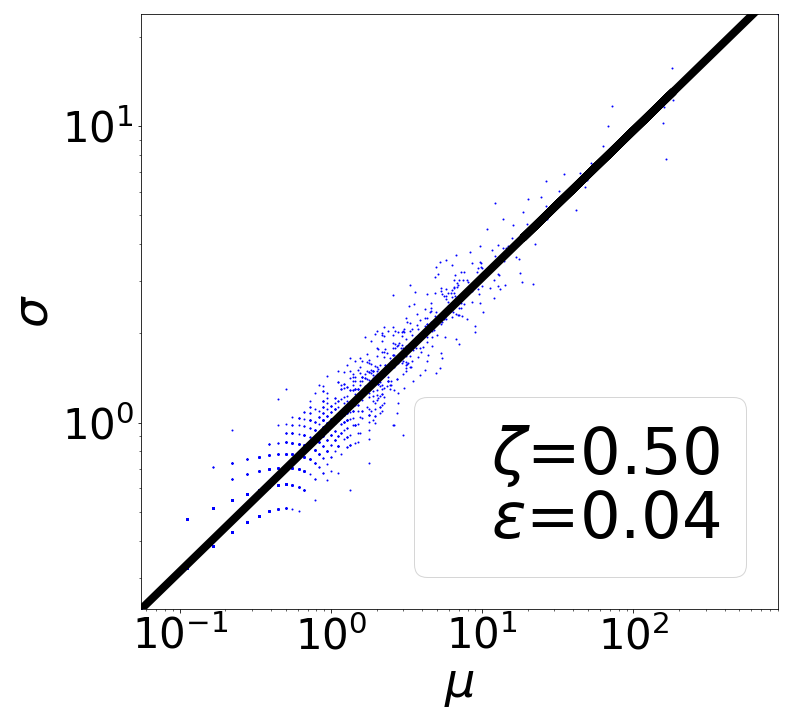}
	\vspace*{-0.5cm}
 \caption{Scaling properties of captions in the COCO image dataset. \label{fig:coco}}
 	\twofig{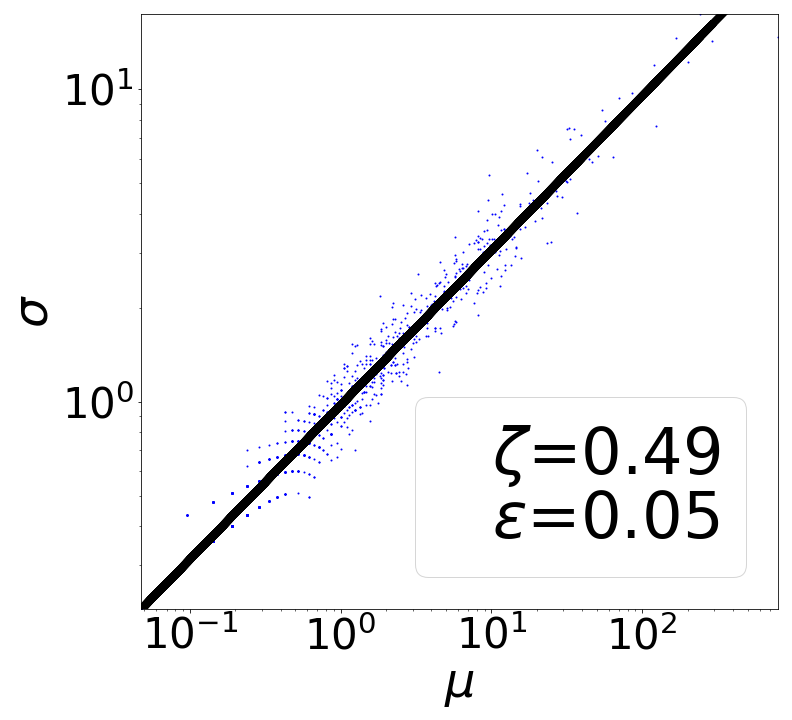}
 \caption{Scaling properties of captions generated from the COCO image dataset by SeqGAN. \label{fig:gan}}
 \end{figure}

\begin{table*}[h]
\footnotesize
\centering
\caption{BLEU scores and perplexity for eval-AWD-LSTM-based evaluation on texts generated from the COCO image dataset by different GAN models: SeqGAN \cite{Yu_2017}, MaliGAN \cite{Che_2017}, RankGAN \cite{LinK_2017}, LeakGAN \cite{Guo_2017}, and TextGAN \cite{Zhang_2017}.}
\label{tab:gan}
\begin{tabular}{|l|c|c|c|c|c|c|c|}
\hline
                & SeqGAN & MaliGAN & RankGAN        & LeakGAN        & TextGAN         & MLE    & ImageCoco \\ \hline
BLEU-2          & 0.92  & 0.89   & \textbf{0.94} & 0.93          & 0.65           & 0.92  & 1.00      \\ \hline
BLEU-3          & 0.75  & 0.70   & 0.80          & \textbf{0.82} & 0.65          & 0.68  & 1.00      \\ \hline
BLEU-4          & 0.53  & 0.48   & 0.60          & \textbf{0.66} & 0.60           & 0.57  & 1.00      \\ \hline
BLEU-5          & 0.35  & 0.31   & 0.41          & 0.47         & \textbf{0.52}  & 0.39  & 1.00      \\ \hline
eval-AWD-LSTM & 179.29 & 272.53  & 132.90         & 146.26         & \textbf{129.93} & 176.34 & 44.17     \\ \hline
\end{tabular}

\end{table*}

\section{Evaluation of GAN Models}
\label{sec:gan}
Finally, we discuss the possibility of evaluating GAN-generated text with the scaling properties. \tabref{tab:coco} lists the scaling properties for the COCO image dataset \cite{Lin_2014}. Because current GAN models for text generation cannot produce long texts, image captions constitute the standard dataset for these GAN models. Because of the dataset used, the GAN models are limited to generating a certain text type (i.e., image captions). In particular, as the length of the text is short, the results are readily expected not to reproduce long memory behavior. Yet, it is worthwhile to test the vocabulary population of the GAN models to understand their capacity.

\figref{fig:coco} and \figref{fig:gan} show Zipf's and Taylor's law graphs for the original dataset and the text generated by SeqGAN \cite{Yu_2017}, respectively. Unlike the other language models, GAN models for text generation had problems reproducing Zipf's law. The tail decay for the generated text was faster than that for the dataset. The vocabulary size of the generated text was only $v(n)=1,822$ words for $n=118,264$ generated words, whereas the original text had a vocabulary size $v(n)=6,095$ for $n=105,933$ words. This result indicates that the GAN model could not produce the infrequent words in the training dataset.

On the other hand, long memory was already absent at the level of the training dataset. The Taylor exponent was $\zeta=0.50$ (\figref{fig:coco}(b)), indicating no memory, which was obviously expected, as the captions were shuffled and two consecutive captions had no relation. Through learning of such training data and production caption by caption, the generated text also had no long memory (\figref{fig:gan}(b)). Indeed, long memory analysis literally requires a model to generate a sufficiently long text to allow further quality evaluation of natural language. 

Nevertheless, other metrics would not provide a better evaluation in this case. \tabref{tab:gan} lists the evaluation metrics of BLEU and perplexity by eval-AWD-LSTM for texts generated using different GAN techniques. The BLEU scores for the GAN models in \tabref{tab:gan} were extracted from \cite{Zhu_2018}. The perplexity scores were computed by using the eval-AWD-LSTM model trained with the COCO image dataset and the hyperparameters for the PTB dataset. The perplexity of AWD-LSTM when trained with that dataset was 65.41.

For both BLEU and perplexity, the results were inconsistent. In terms of BLEU, the best-performing GAN model varied among RankGAN with BLEU-2, LeakGAN with BLEU-3 and BLEU4, and TextGAN with BLEU-5. In contrast, TextGAN was the best model in terms of eval-AWD-LSTM. In addition to these metrics, the negative log-likelihood of the PCFG was also not effective in evaluating the GAN models in \cite{Zhu_2018}.

Although rigid quantitative evaluation is necessary for comparing GAN models, the existing evaluation metrics are not sufficiently reliable. Therefore, further study of evaluation metrics is necessary. The Taylor exponent may play a role in such studies when GAN-based models become able to produce longer texts.

\section{Conclusion}
\label{sec:conc}
In this article, we have investigated the scaling properties of computational models of natural language and analyzed whether these metrics could serve for assessing the models. The scaling properties quantify the vocabulary population and long memory behavior, which are universal qualities of natural language text. These metrics are applicable to any model, even those for which the perplexity is not measurable or a reference is not available. We tested $n$-gram language models, a grammatical model, mathematical models, neural language models, and GAN models for text generation. Among the five scaling properties introduced, the exponent of Taylor's law showed the most reasonable behavior. It had the clearest correlation with the perplexity of the $n$-gram and neural language models.

Our analysis demonstrated that RNNs with a gating mechanism (LSTM, GRU, and QRNNs) are the first computational models of natural language that have the capacity to reproduce the long memory in natural language text. No other models tested in our experiment reproduced the scaling properties of long memory. The LSTM models were the best among the neural language models, as their long memory behavior was closer to that of the original text as compared to the GRU and QRNN models. Yet, even the LSTM language models could not entirely recover long memory, including the exponents of the scaling properties. This observation confirms the gap between natural language text and language models and suggests corresponding room for improvement. Our future work will include investigating other scaling properties that could serve for evaluating language models.
\newpage
\starttwocolumn

\bibliographystyle{compling}
\bibliography{cl18.bib}

\onecolumn
\renewcommand{\thefigure}{A\arabic{figure}}
\setcounter{figure}{0}

\newpage

\section*{Appendix A: Scaling properties of natural language}
This section presents the figures for the scaling properties of dataset appeared in this paper. The presence of the scaling properties is robust to the genre and the language of the text.
\label{sec:moreexamples}

\begin{figure}[h]
	\tinyallfig{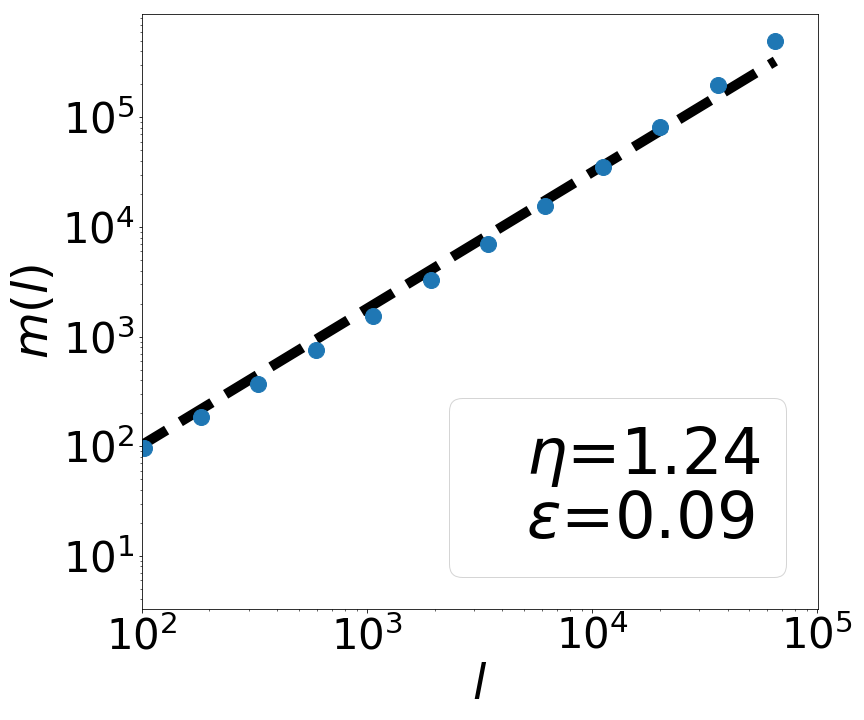}
	\vspace*{-0.5cm}
	\caption{Scaling properties of the collected works of Shakespeare}
	\label{fig:shakespeare}
\end{figure}

\begin{figure*}[h]
  \tinyallfig{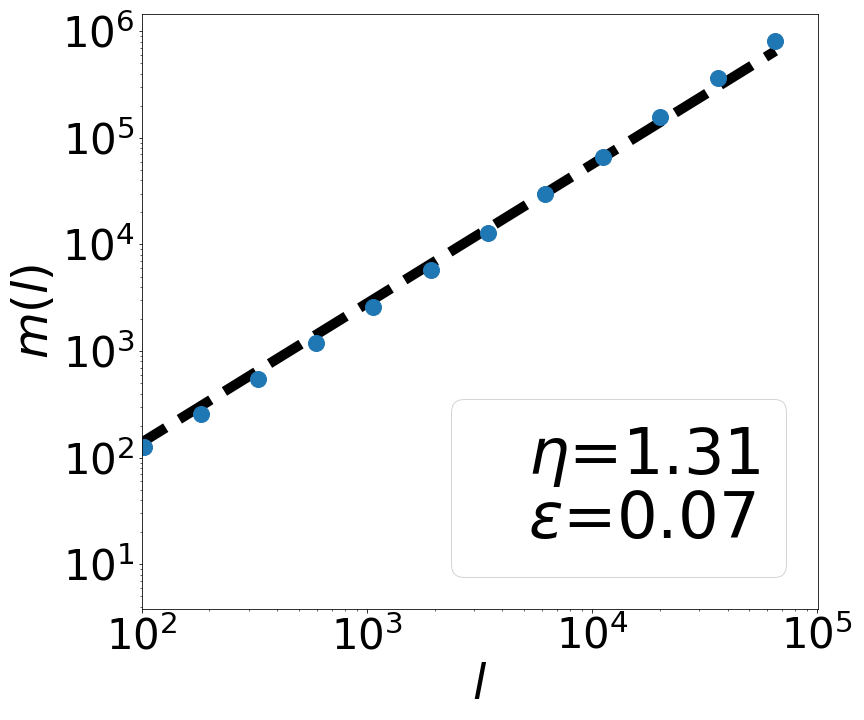}
  \vspace*{-0.5cm}
\caption{Scaling properties of Hong Lou Meng.}
\label{fig:koromu}
\end{figure*}

 \begin{figure*}[h]
   \tinyallfig{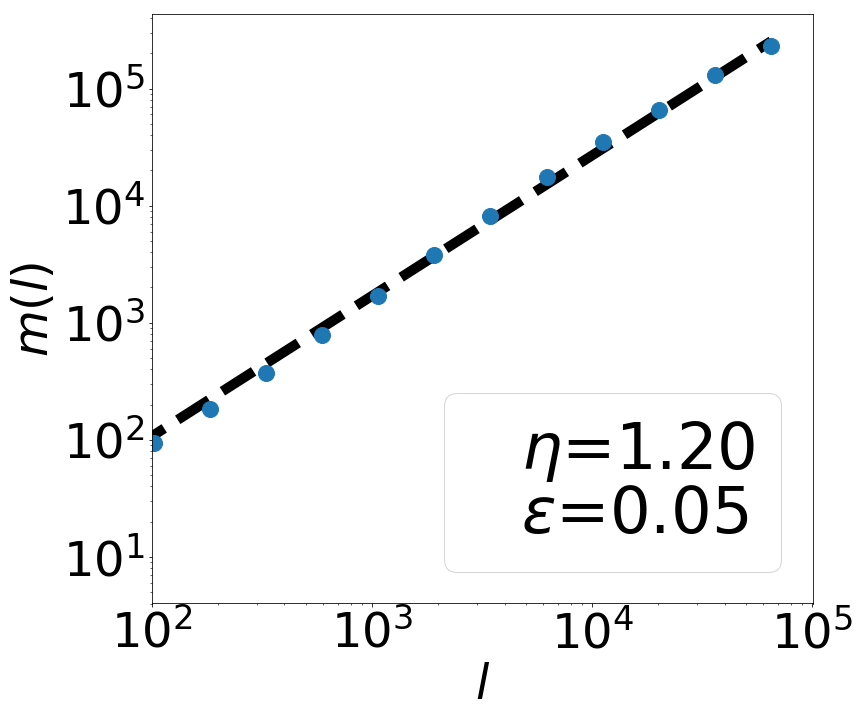}
   \vspace*{-0.5cm}
 \caption{Scaling properties of the Penn-Treebank (original)}
 \end{figure*}

 \begin{figure*}[h]
   \tinyallfig{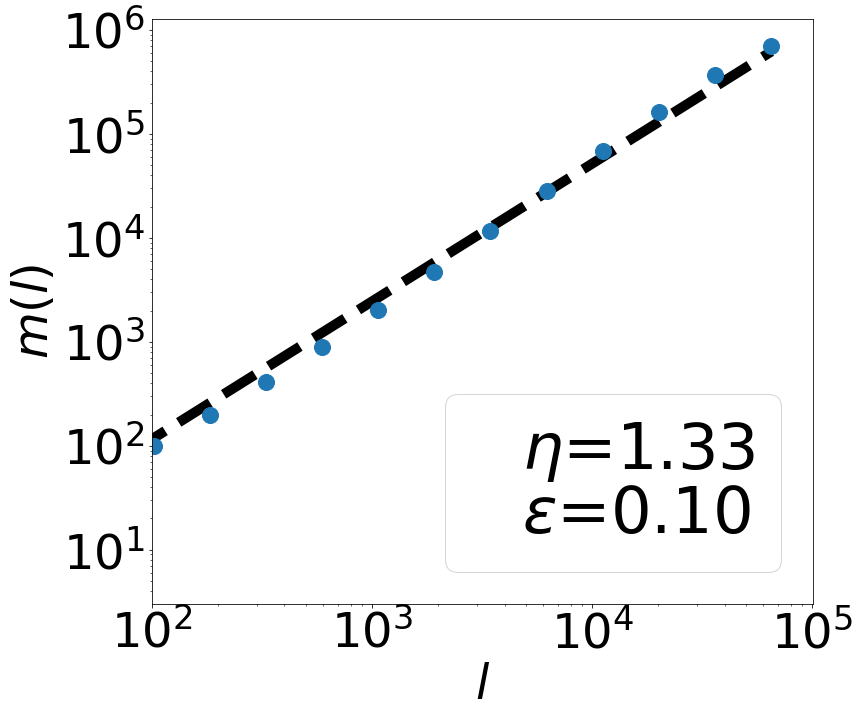}
   \vspace*{-0.5cm}
 \caption{Scaling properties of Wikitext-2 (original)}
 \end{figure*}
 
  \begin{figure*}[h]
   \tinyallfig{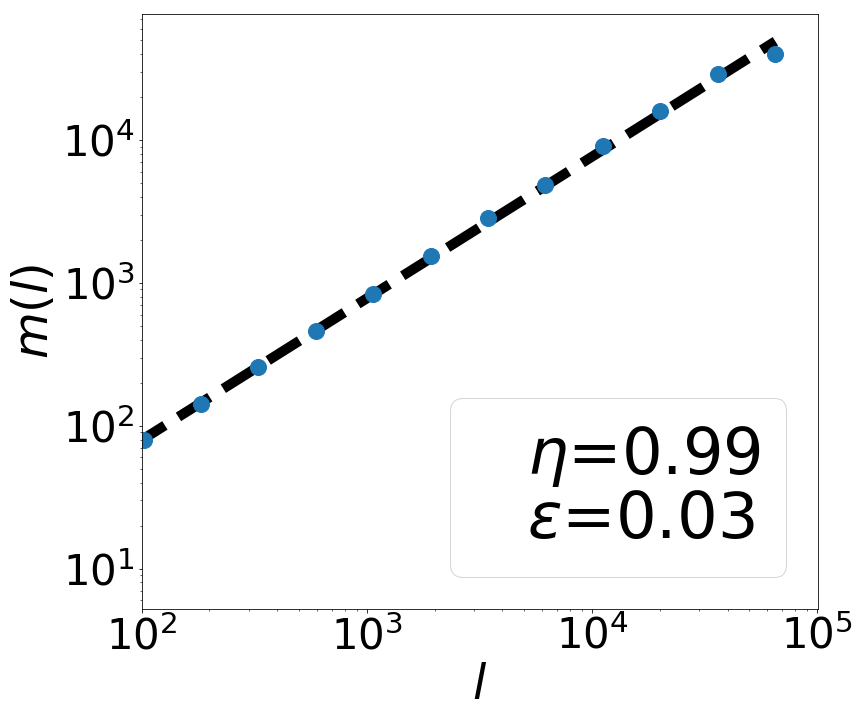}
   \vspace*{-0.5cm}
 \caption{Scaling properties of captions in COCO image dataset}
 \end{figure*}

\newpage

\section*{Appendix B: Scaling properties of language model}
\label{sec:lm_examples}
This section presents the figures for the scaling properties of language models of WT2 in this paper.

 \begin{figure*}[h]
   \tinyallfig{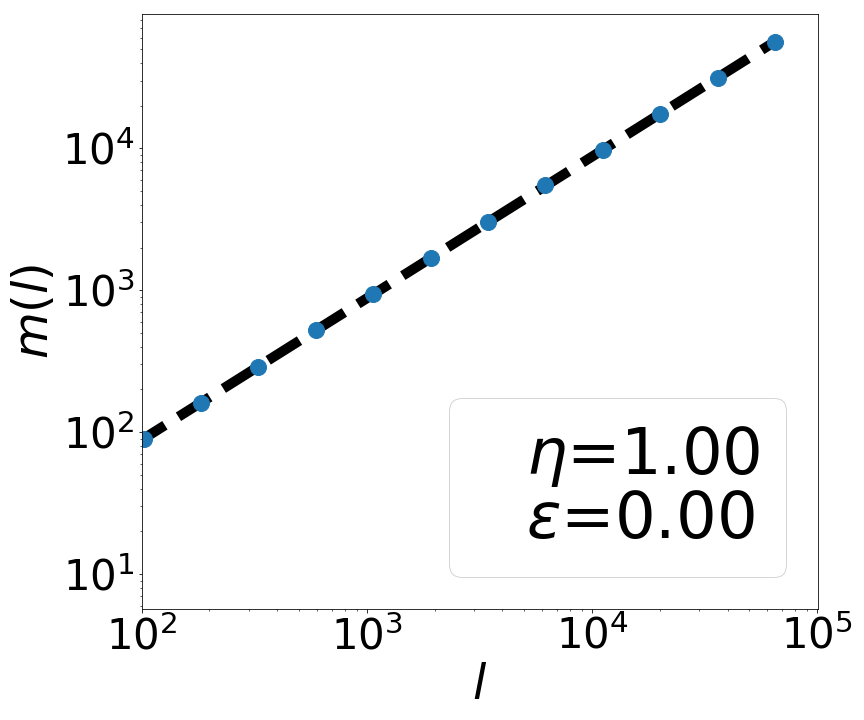}
   \vspace*{-0.5cm}
 \caption{Scaling properties of 3-gram language model}
 \end{figure*}

 \begin{figure*}[h]
   \tinyallfig{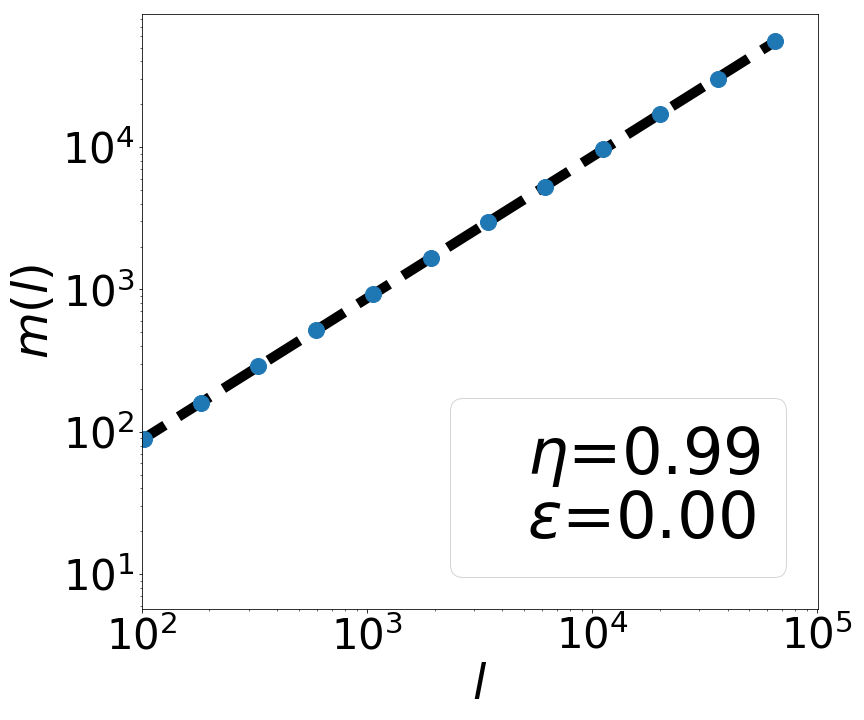}
   \vspace*{-0.5cm}
 \caption{Scaling properties of 5-gram language model}
 \end{figure*}

 \begin{figure*}[h]
   \tinyallfig{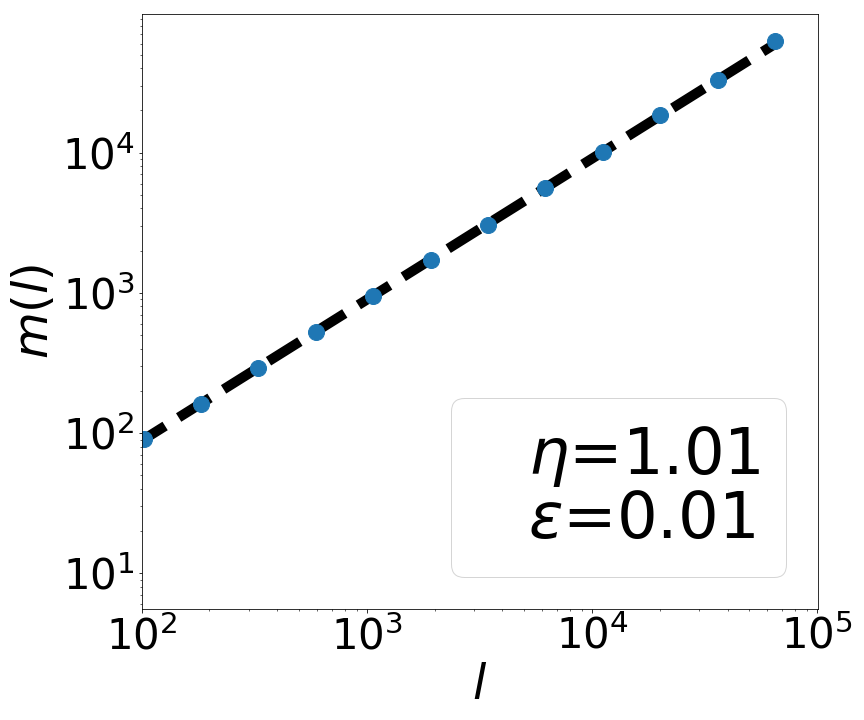}
   \vspace*{-0.5cm}
 \caption{Scaling properties of linear interpolation $n$-gram language model }
 \end{figure*}

 \begin{figure*}[h]
   \tinyallfig{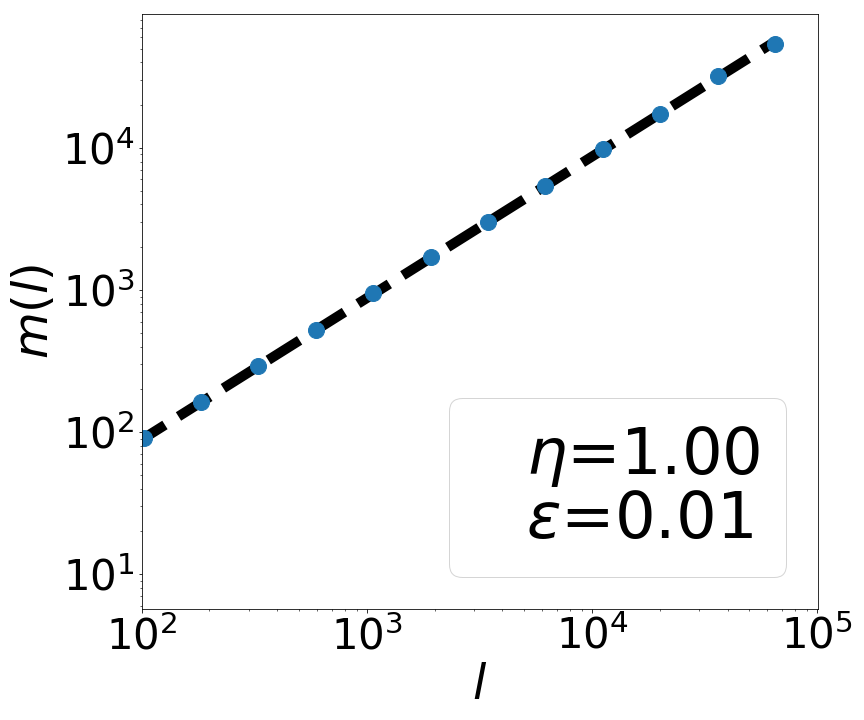}
   \vspace*{-0.5cm}
 \caption{Scaling properties of the Katz backoff 3-gram language model}
 \end{figure*}

 \begin{figure*}[h]
   \tinyallfig{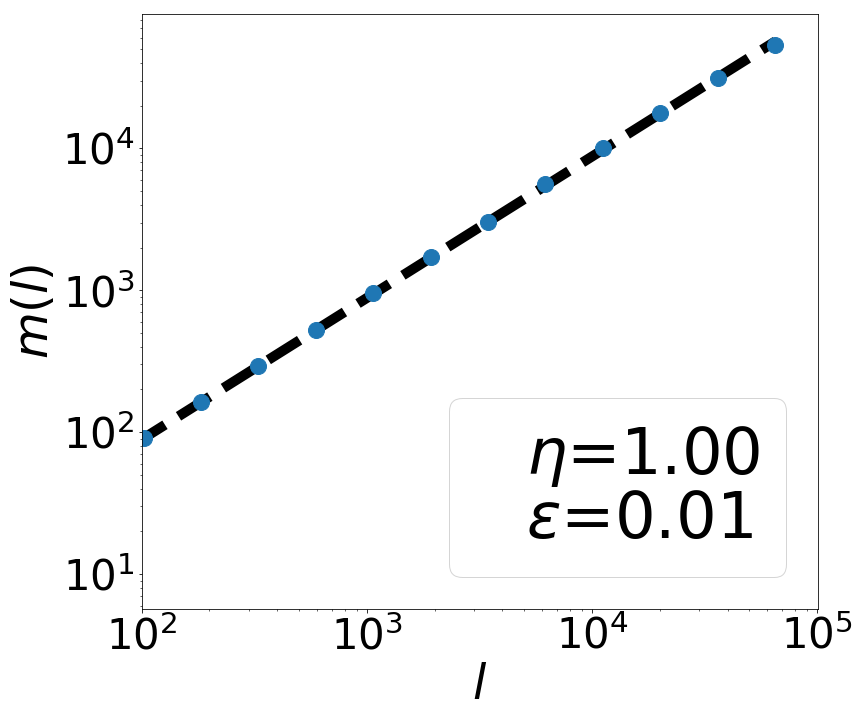}
   \vspace*{-0.5cm}
 \caption{Scaling properties of Katz backoff 5-gram language model}
 \end{figure*}
 
  \begin{figure*}[h]
   \tinyallfig{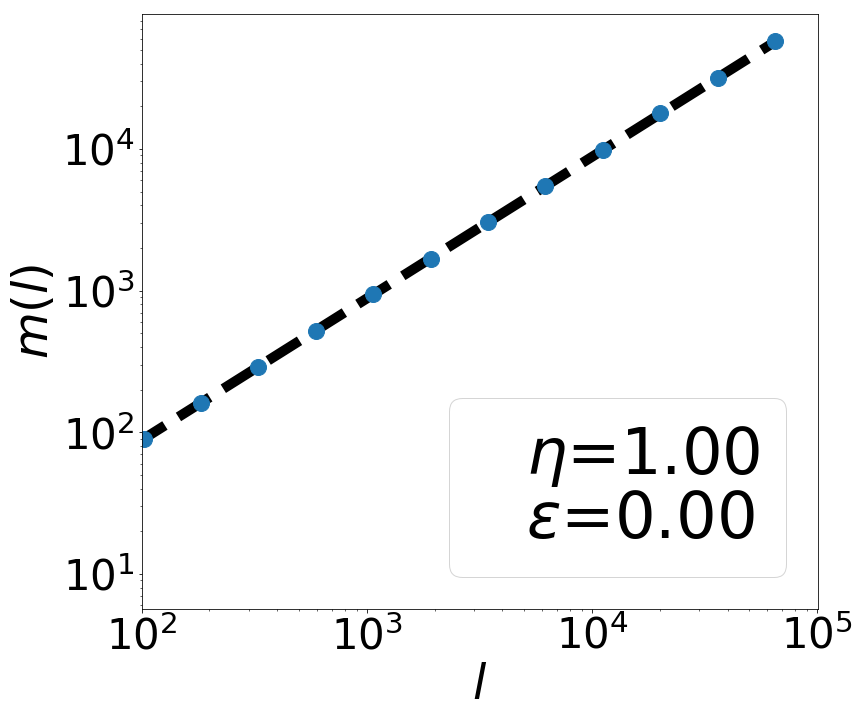}
   \vspace*{-0.5cm}
 \caption{Scaling properties of Kneser-Ney 3-gram language model}
 \end{figure*}
 
  \begin{figure*}[h]
   \tinyallfig{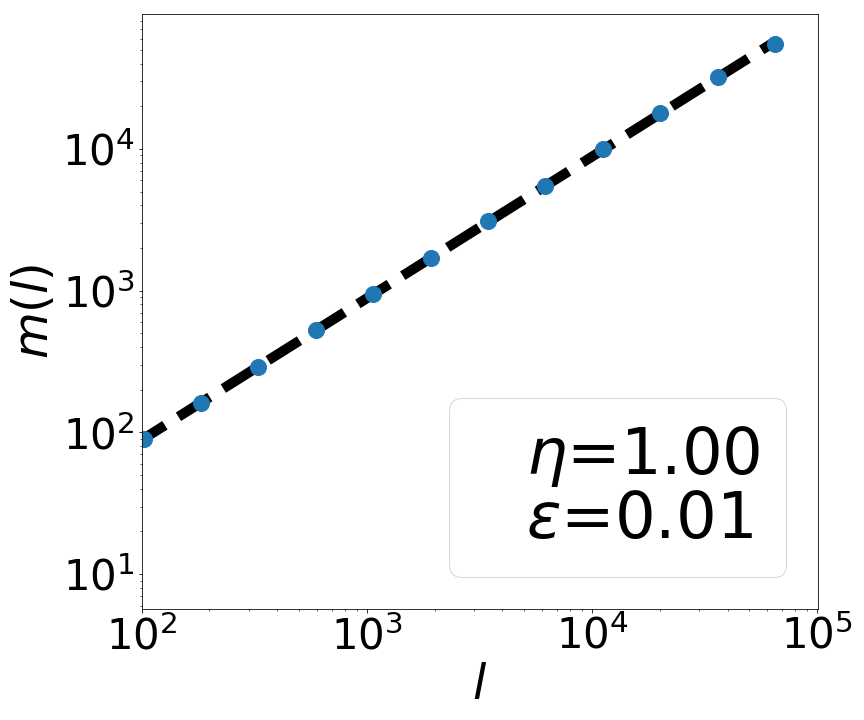}
   \vspace*{-0.5cm}
 \caption{Scaling properties of Kneser-Ney 5-gram language model}
 \end{figure*}
 
   \begin{figure*}[h]
   \tinyallfig{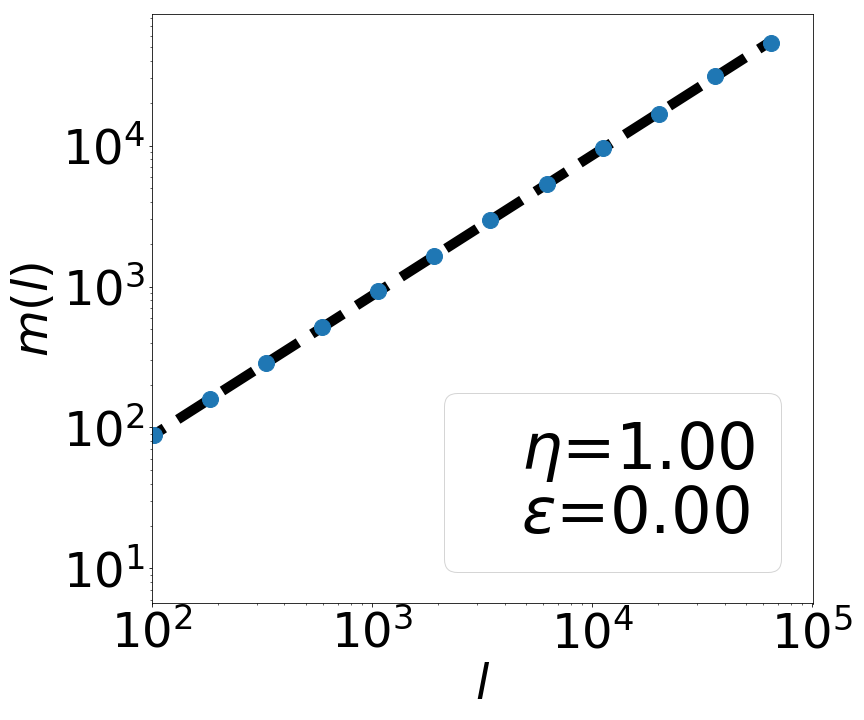}
   \vspace*{-0.5cm}
 \caption{Scaling properties of hierarchical Pitman-Yor language model}
 \end{figure*}
 
   \begin{figure*}[h]
   \tinyallfig{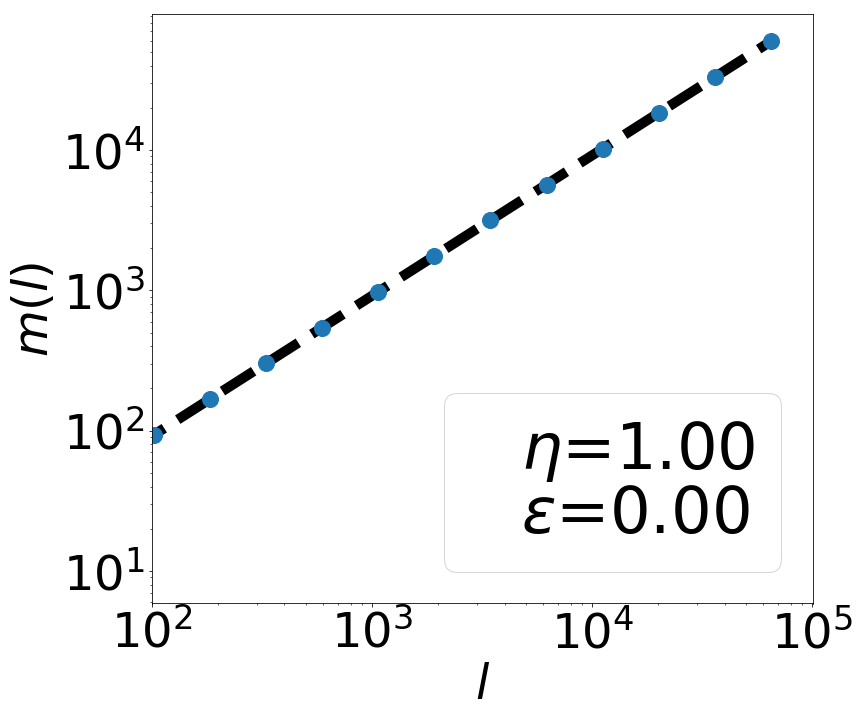}
   \vspace*{-0.5cm}
 \caption{Scaling properties of the PCFG constructed from PTB dataset}
 \end{figure*}
 
\begin{figure*}[h]
   \tinyallfigwordNoEbeling{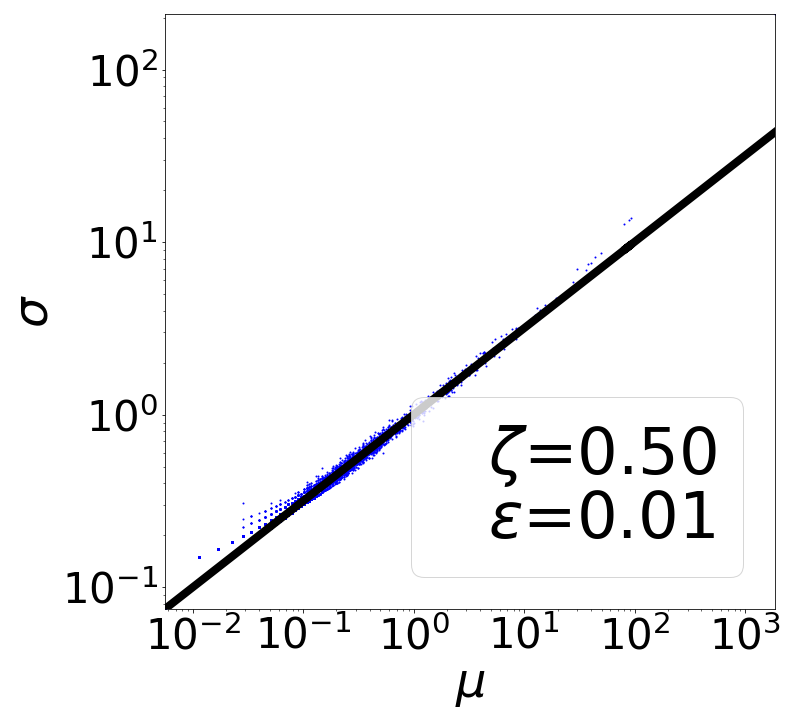}
   \vspace*{-0.5cm}
 \caption{Scaling properties of the Simon process. The figure of Ebeling method does not appear because of the inappropriateness of the application.}
\label{fig-appendix:simon}
\end{figure*}
 
     \begin{figure*}[h]
   \tinyallfigwordNoEbeling{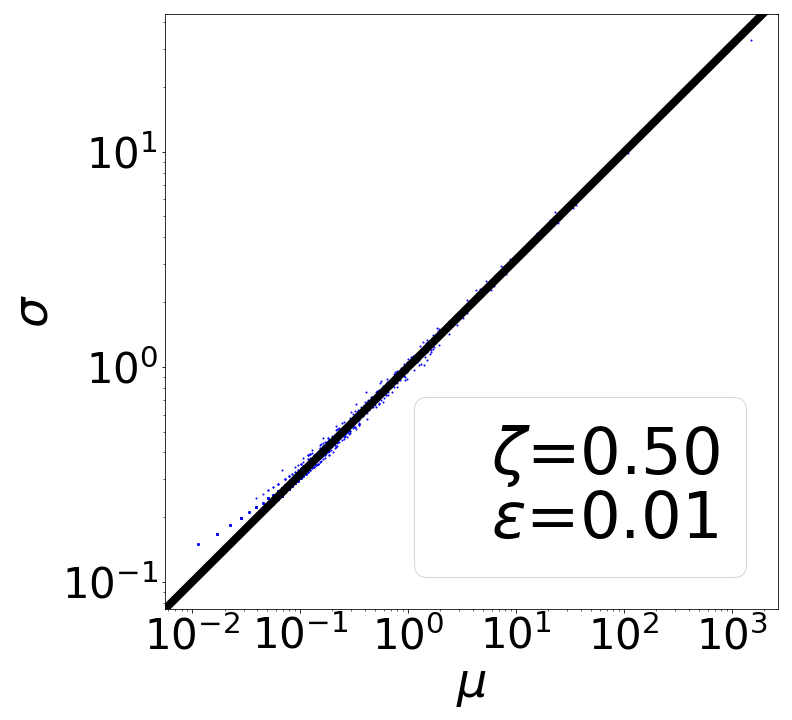}
   \vspace*{-0.5cm}
 \caption{Scaling properties of Pitman-Yor process. The figure of Ebeling method does not appear because of the inappropriateness of the application.} 
 \end{figure*}

\begin{figure*}[h]
   \tinyallfig{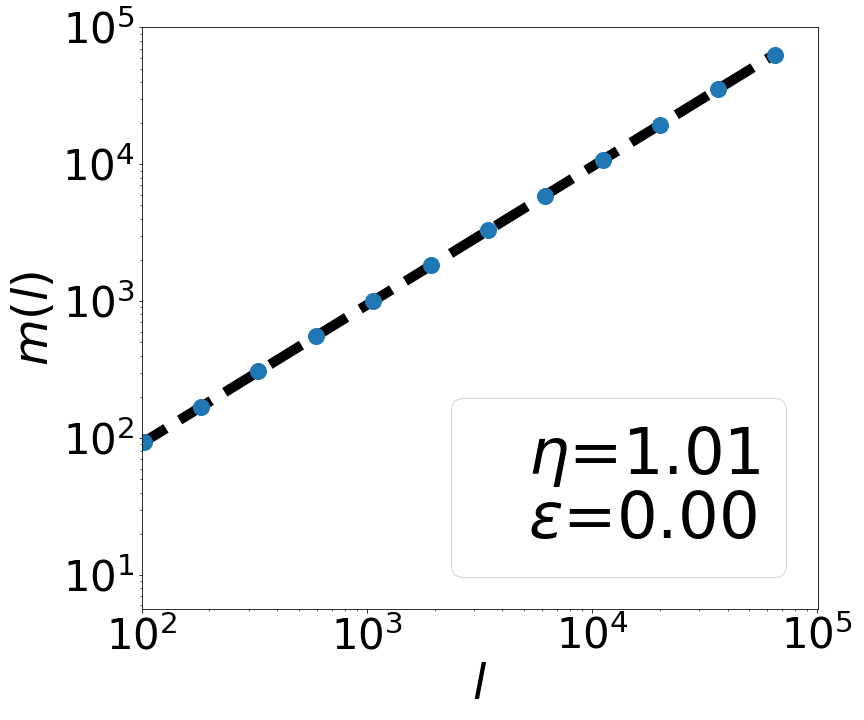}
   \vspace*{-0.5cm}
 \caption{Scaling properties of Simple RNN language model}
 \end{figure*}
 
\begin{figure*}[h]
	\tinyallfig{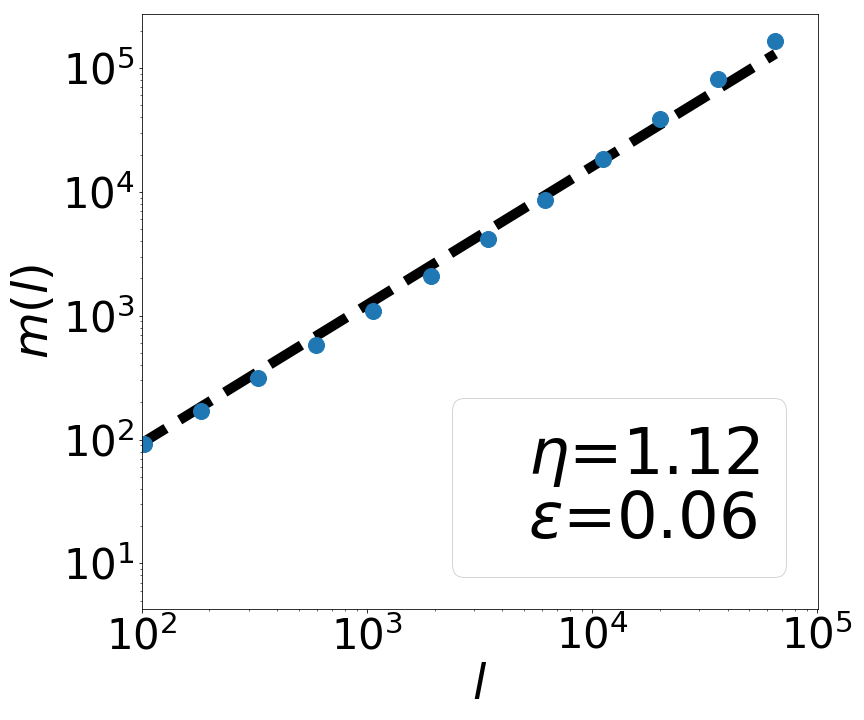}
	\vspace*{-0.5cm}
	\caption{Scaling properties of GRU language model}
\end{figure*}

\begin{figure*}[h]
	\tinyallfig{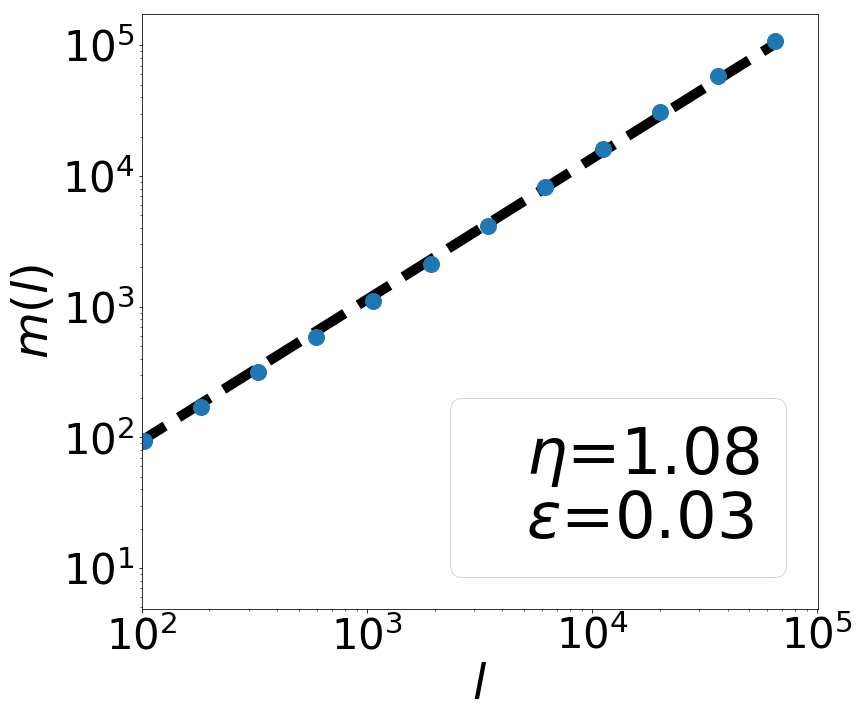}
	\vspace*{-0.5cm}
	\caption{Scaling properties of QRNN language model}
\end{figure*}

\begin{figure*}[h]
   \tinyallfig{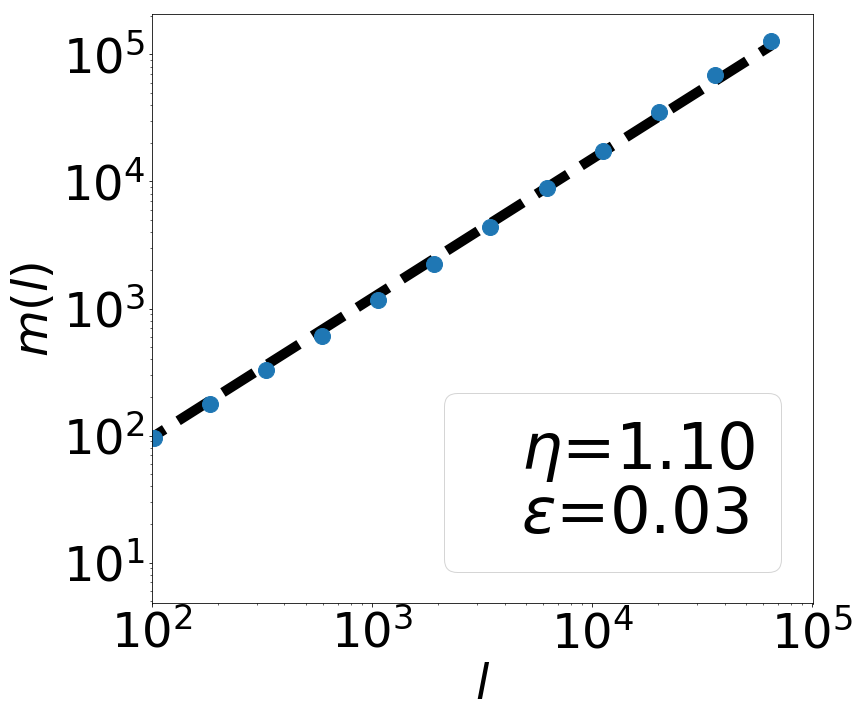}
   \vspace*{-0.5cm}
 \caption{Scaling properties of LSTM without regularization language model}
 \end{figure*}
 
   \begin{figure*}[h]
   \tinyallfig{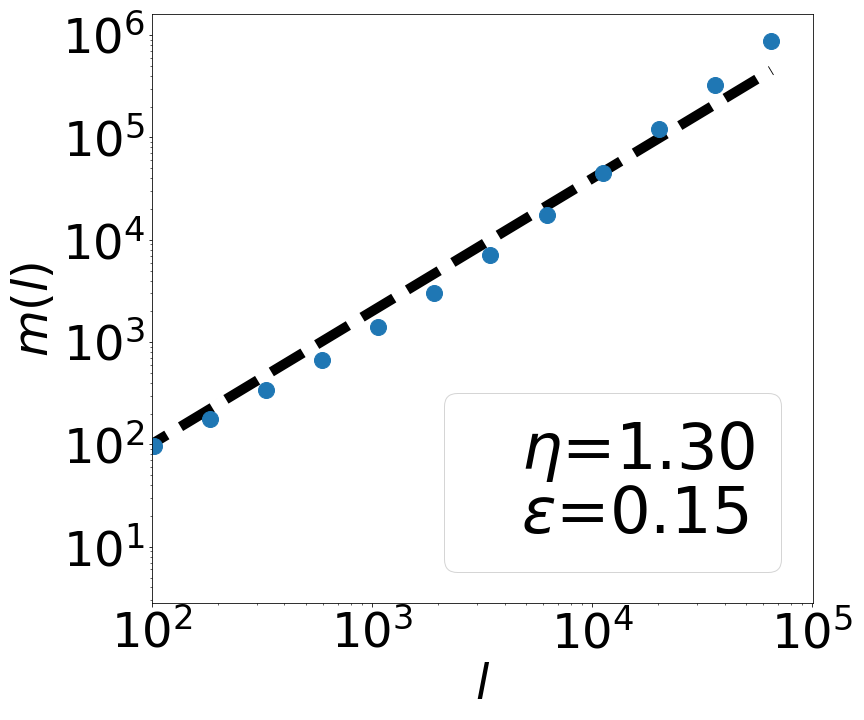}
   \vspace*{-0.5cm}
 \caption{Scaling properties of AWD-LSTM}
 \end{figure*}
 
   \begin{figure*}[h]
   \tinyallfig{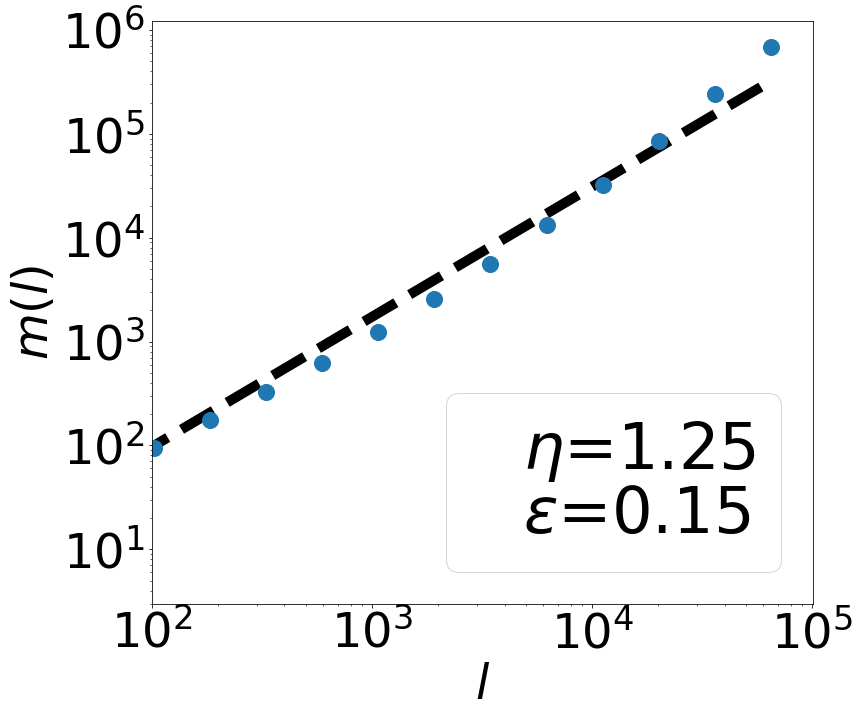}
   \vspace*{-0.5cm}
 \caption{Scaling properties of AWD-LSTM-Simon}
 \end{figure*}
 
    \begin{figure*}[h]
   \tinyallfig{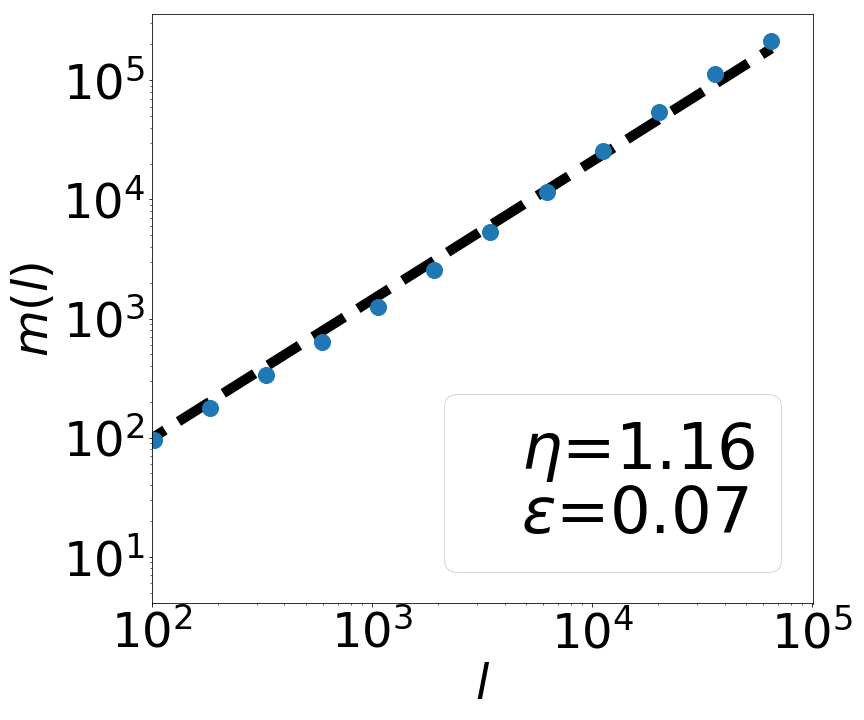}
   \vspace*{-0.5cm}
 \caption{Scaling properties of AWD-LSTM-MoS}
 \end{figure*}
 
     \begin{figure*}[h]
   \tinyallfig{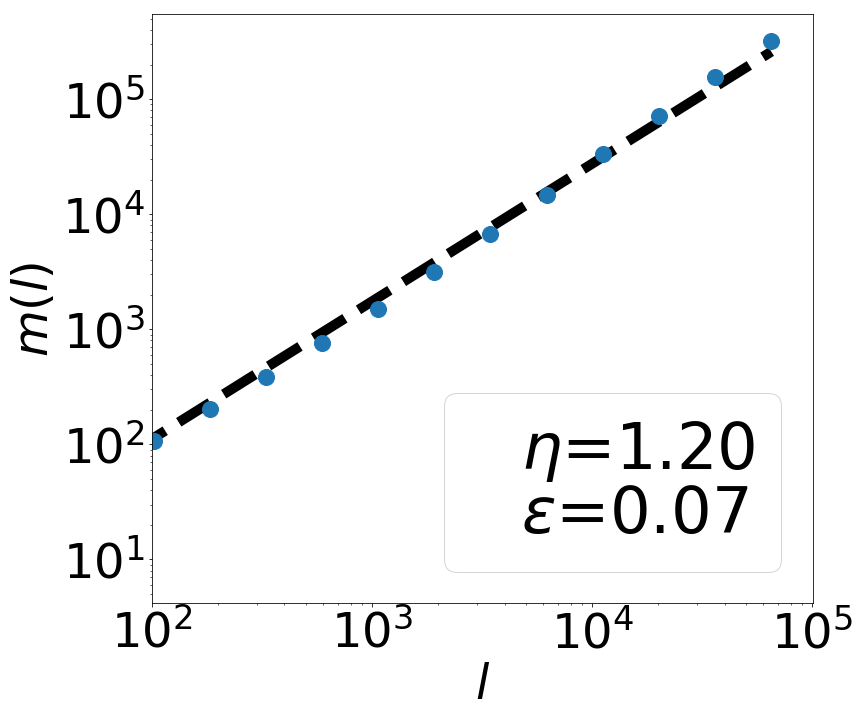}
   \vspace*{-0.5cm}
 \caption{Scaling properties of AWD-LSTM-MoS-Cache}
 \end{figure*}
 
 \begin{figure*}[h]
\tinyallfig{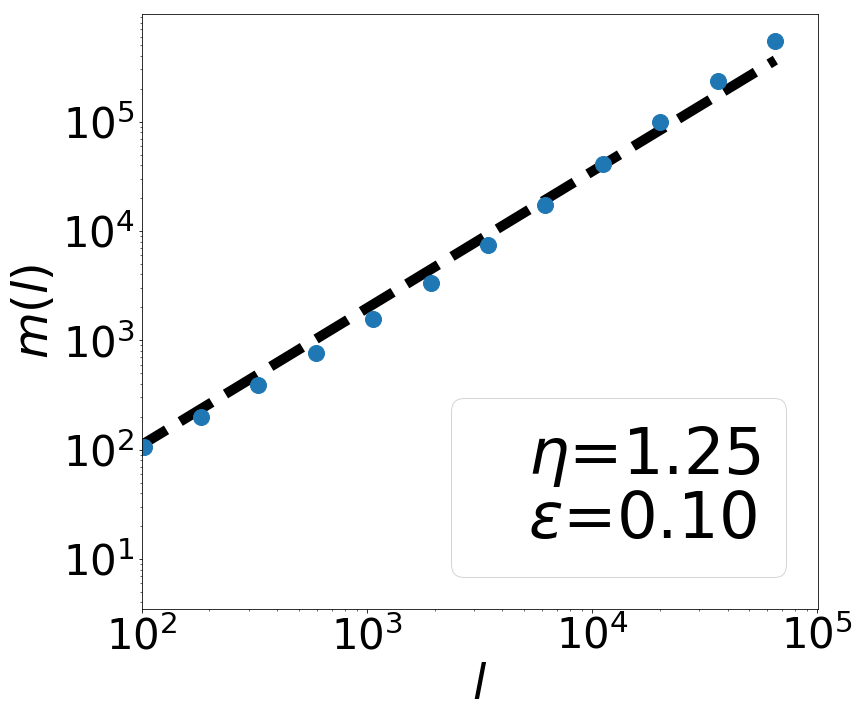}
   \vspace*{-0.5cm}
 \caption{Scaling properties of AWD-LSTM-Cache}
 \end{figure*}
 
  \begin{figure*}[h]
\tinyallfig{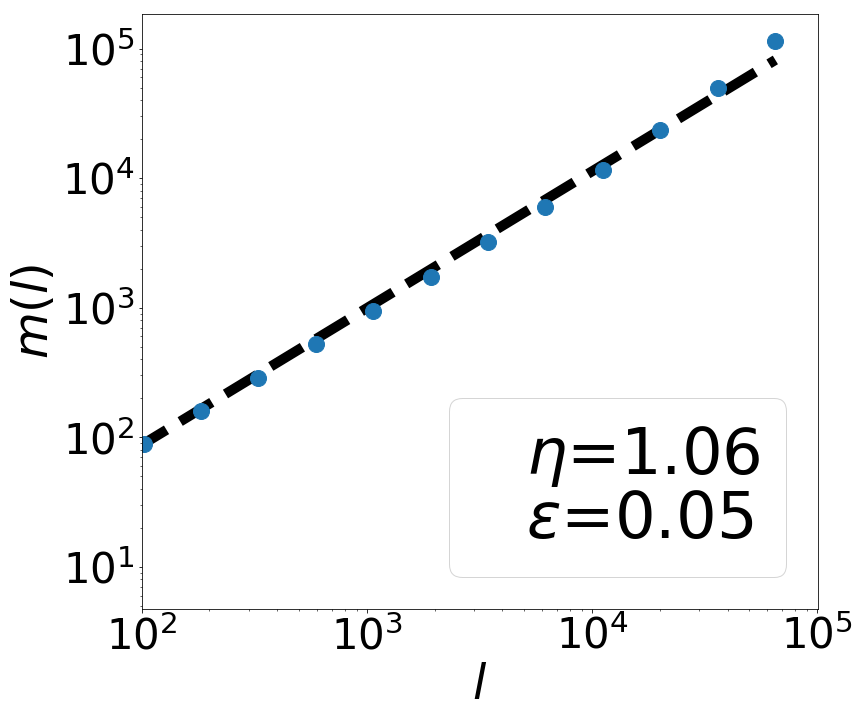}
\vspace*{-0.5cm}
 \caption{Scaling properties of LSTM without regularization for character-level modeling}
 \end{figure*}

\begin{figure*}[h]
\tinyallfig{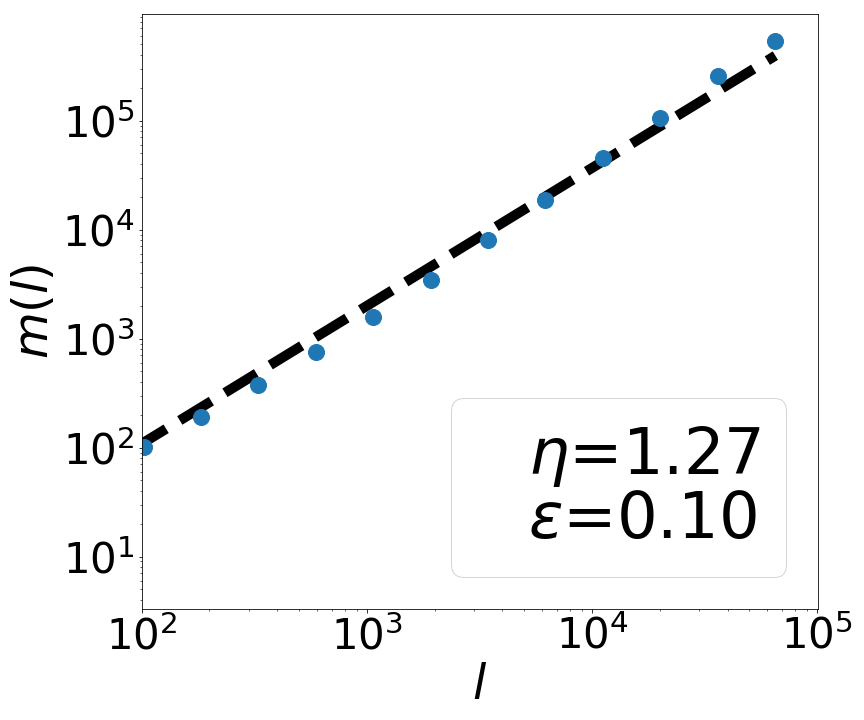}
\vspace*{-0.5cm}
 \caption{Scaling properties of AWD-LSTM for character-level modeling}
 \end{figure*}

 \begin{figure*}[h]
   \tinyallfig{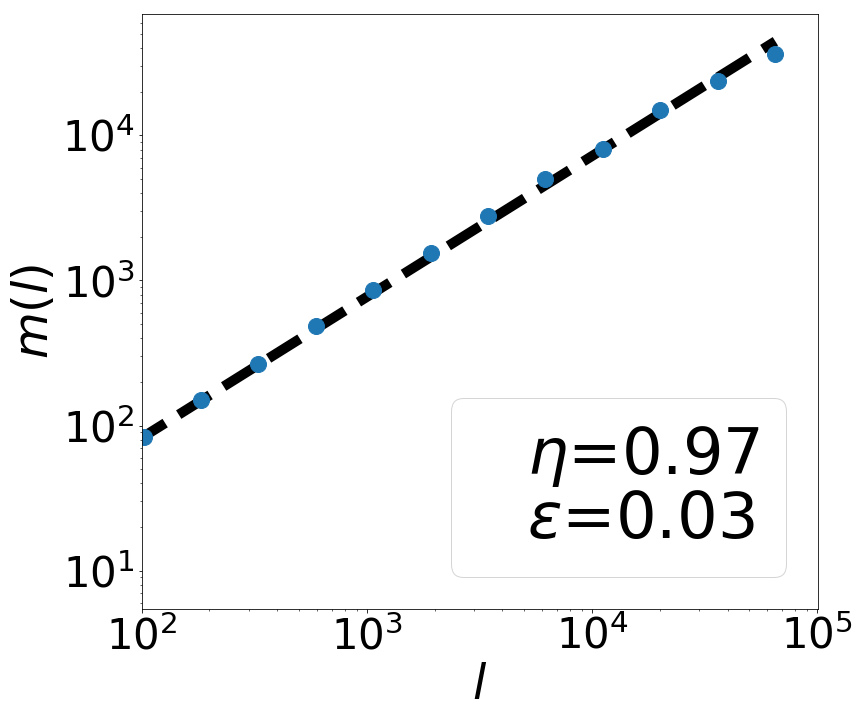}
   \vspace*{-0.5cm}
 \caption{Scaling properties of the Seq-GAN (the model learns COCO image dataset)}
 \end{figure*}

\end{document}